%% file: cas-sc-template.tex
\documentclass[a4paper]{cas-sc}

\usepackage[numbers]{natbib}
\usepackage{float}
\usepackage{placeins}
\def\tsc#1{\csdef{#1}{\textsc{\lowercase{#1}}\xspace}}
\tsc{WGM}
\tsc{QE}
\tsc{EP}
\tsc{PMS}
\tsc{BEC}
\tsc{DE}
\begin{document}
\let\WriteBookmarks\relax
\def\floatpagepagefraction{1}
\def\textpagefraction{.001}

\shorttitle{Propagation and Sharing of Medical Norms}

\shortauthors{Chao Li,Sergey Kovalchuk et~al.}

\title [mode = title]{Multi-Agent Norm Perception and Induction in Distributed Healthcare}                      
\tnotetext[1]{This research is financially supported by The Russian Science Foundation, Agreement 24-11-00272.}
% \tnotetext[2]{}
\author[1]{Chao Li}[orcid=0009-0000-1233-3281]
\cormark[1]
\ead{316325@niuitmo.ru}
% \credit{Conceptualization of this study, Methodology, Writing - Original draft preparation}

\author[2]{Olga Petruchik}[orcid=0000-0001-6855-9829]
% \fnmark[3]
\ead{petru4ikolga2017@gmail.com}
% \credit{Data curation, Investigation}

\author[1]{Elizaveta Grishanina}
% \fnmark[4]
\ead{eogrishanina@niuitmo.ru}
% \credit{Formal analysis, Resources}

\author[1]{Sergey Kovalchuk}[orcid=0000-0001-8828-4615]
\cormark[2]
\ead{kovalchuk@itmo.ru}
% \credit{Supervision, Validation, Writing - Review \& Editing}

\affiliation[1]{organization={ITMO University},
    addressline={}, 
    city={Saint Petersburg},
    postcode={197101}, 
    country={Russia}}

\affiliation[2]{organization={Taganrog City Clinical Emergency Hospital},
    addressline={}, 
    city={Taganrog},
    postcode={}, 
    country={Russia}}
\cortext[1]{Corresponding author}
\cortext[2]{Principal corresponding author}
% \fntext[1]{This is the first author footnote.}
% \fntext[2]{This is the second author footnote.}
% \fntext[3]{This is the third author footnote.}
% \fntext[4]{This is the fourth author footnote.}

% \nonumnote{This note has no numbers. In this work we demonstrate $a_b$
%   the formation Y\_1 of a new type of polariton on the interface
%   between a cuprous oxide slab and a polystyrene micro-sphere placed
\begin{abstract}
This paper presents a Multi-Agent Norm Perception and Induction Learning Model aimed at facilitating the integration of autonomous agent systems into distributed healthcare environments through dynamic interaction processes. The nature of the medical norm system and its sharing channels necessitates distinct approaches for Multi-Agent Systems to learn two types of norms. Building on this foundation, the model enables agents to simultaneously learn descriptive norms, which capture collective tendencies, and prescriptive norms, which dictate ideal behaviors. Through parameterized mixed probability density models and practice-enhanced Markov games, the multi-agent system perceives descriptive norms in dynamic interactions and captures emergent prescriptive norms. We conducted experiments using a dataset from a neurological medical center spanning from 2016 to 2020. 
In the model’s descriptive norm-sharing experiment, within small-scale medical communities, each agent's Subjective Individual Norm Perception (SINP) achieved a stable $KL$ divergence of below 0.1 with the Objective Collective Norm (OBJ) after participating in medical practice-sharing activities, indicating that the model can perceive the collective medical norms representing the current best clinical practices in the environment. In medium-scale medical communities, the relationship between the number of agents and the norm convergence times followed different types of heavy-tailed distributions. In large-scale medical communities, the system's convergence rate exhibited fluctuations; however, overall, it decreased linearly as the number of agents increased. The Practice Shannon Diversity Index of the medical community, consisting of 10 neurologists in the dataset, gradually converged over 5 years to a value that better aligns with the real diagnostic practices of the neurological medical center. This indicates that the model, through long-term learning and sharing processes, can progressively reflect the actual diagnostic trends and collective behavioral tendencies within the medical community. In the experiment where multiple agents infer prescriptive norms within a dynamic healthcare environment, the agents effectively learned the key clinical protocols $\mathcal{N}_1$ and $\mathcal{N}_2$ within the norm space $\mathcal{H}$, which also included 30 control norms, without developing high belief in invalid norms. Furthermore, the agents’ belief update process was relatively smooth, avoiding any discontinuous stepwise updates.

\end{abstract}

% \includegraphics{figs/grabs.pdf}

% \begin{highlights}
% \item Research highlights item 1
% \item Research highlights item 2
% \item Research highlights item 3
% \end{highlights}

% Keywords

\maketitle

\section{Introduction}

Although the majority of studies have focused on a head-to-head comparison of AI with humans~\cite{Liu2019ACO}, real-life medical practice is more likely to involve human-in-the-loop setups, where humans actively collaborate with AI systems and provide oversight~\cite{Chen2019AnAR,Patel2019HumanmachinePW}. There are still open questions about exactly how AI assistance affects human performance~\cite{Rajpurkar2022AIIH}. Recent studies have begun to explore such collaborative setups between AI and humans from multiple perspectives~\cite{Patel2019HumanmachinePW,Sim2019DeepCN,Tschandl2020HumancomputerCF,Kiani2020ImpactOA}.

In a highly distributed and fragmented environment like healthcare~\cite{Kovalchuk2017SimulationOP,Rajpurkar2022AIIH,Guleva2020EmergingCI}, characterized by strong local autonomy and dynamism~\cite{Schoenherr2021TrustAE,Grimson2001SharingHR,Funkner2021CitywideQO}, the key to addressing collaboration between AI and humans lies in the ability of autonomous agents to understand and adhere to the norms among healthcare professionals (HCPs).

The notion of norms is applied in various settings to describe the consistency of thought or behavior within a reference group, typically aiming to understand how these patterns influence the members of those groups~\cite{Horne2020NormsAI,Miller2016ChangingNT,Morris2015NormologyII,Cotterill2019TheIO}. Ethicists and philosophers typically conceptualize norms as being strictly divided into descriptive norms and prescriptive norms~\cite{Donaldson1994TowardAU,Goldstein2007UsingSN,Morris2015NormologyII}. Norms play a critical role in facilitating cooperation and collaboration within groups, communities, and societies. They serve as indispensable mechanisms that uphold collective action~\cite{Nyborg2016SocialNA,DAgostino1996ContemporaryAT,Warner2016AnEP}. Prescriptive norms set forth how people ought to behave~\cite{Kalantari2013OnTL}. Prescriptive norms can be transmitted via messaging and reinforced by the constraints imposed by individuals' social roles or positions~\cite{Spicer2004DoesNC}. Descriptive norms capture the observed or communicated tendencies of behavior or thought within a group, which can be conveyed to others through observation or the transmission of informal or formal information~\cite{Scharding2023WhenAN,Kalantari2013OnTL}.

Most of the medical norms are for human doctors, and are written in natural language with weak formalization and references to doctors' expertise. Enabling autonomous agent systems to learn and perceive the norms among healthcare professionals is both intriguing and essential for their integration into real-world healthcare environments. Considering the general properties of norms and the inherent distributed nature of the healthcare environment, we propose a Multi-Agent Norm Perception and Induction Learning Model that enables autonomous agent systems to simultaneously learn both descriptive and prescriptive medical norms in real-world healthcare scenarios.

Placing interaction at the core of descriptive norms conceptualizes them as processes of individual interactions rather than static end products~\cite{Mirski2021ConventionalMA,Dingemanse2023BeyondSA,Smith2007AgentBasedMA}. This conceptualization serves as our general approach to modeling descriptive norms in distributed healthcare scenarios. We model descriptive norms as agents' subjective perceptions of collective norms, where these subjective perceptions represent a generalized abstraction of descriptive norms. This perception represents a form of Theory of Mind, involving the ability to infer others' mental states based on the information conveyed~\cite{Premack1978DoesTC}. We model the perception of norms by individual agents towards the collective using a parameterized weighted sum of mixed probability density functions. The parameters are updated based on prior norm perceptions from the EHR dataset and the ongoing interactions among multiple agents. This parameterized probability density function model effectively represents the descriptive norms among doctors that agents gradually learn during their interactions.

The modeling of prescriptive norms in healthcare has different characteristics. Federal and regional protocols are based on clinical recommendations derived from reliable evidence. These protocols specify the most effective interventions and treatments for specific groups of patients. However, the implementation of these protocols varies across hospitals, each with its unique real-world circumstances~\cite{Kovalchuk2017SimulationOP}. Therefore, we focus on the emergence and convergence of prescriptive norms within the healthcare environment under such dynamic conditions.

Shared and distributed prescriptive norms are learned and sustained in a decentralized and emergent manner~\cite{MorrisMartin2019NormEI,Hadfield2012WhatIL}. Some models, based on the frameworks of joint intentionality in human cooperation~\cite{Tang2020BootstrappingAI} and shared agency~\cite{Tuomela1995TheIO}, allow agents to consider the compliance of the norm while pursuing individual interests~\cite{Oldenburg2024LearningAS}. Agents infer and update their understanding of the norms that best explain collective behavior by analyzing apparent violations of self-interest.

However, contrary to the strict requirements of the healthcare environment, the approximate rational Bayesian rule induction in these methods can lead agents to learn "silly" prescriptive norms within the norm space. To address this, we have refined the Bayesian approach used by agents to induce rules from noisy action data and introduced external regulation by allowing agents to practice and verify norms within the healthcare environment. Additionally, we have incorporated momentum-like mechanisms and adaptive learning rates into the agents’ belief updating process~\cite{Duchi2011AdaptiveSM, Rumelhart1986LearningRB,Kingma2014AdamAM}. The updated model prevents convergence to local optima caused by the internal complexity of the probabilistic model, while simultaneously smoothing the agents’ norm update process, eliminating the discontinuous stepwise changes observed in belief updates.

Building on the distributed nature of healthcare, we developed a model that enables autonomous agent systems to integrate into healthcare environments through dynamic interactions. This approach allows the agents to learn medical norms while taking into account the discrete nature, complexity, and distributed characteristics of the healthcare domain~\cite{Kovalchuk2017SimulationOP,Funkner2021CitywideQO,Li2024ModellingOP}.
Statement of Significance is in the table below.
\begin{table}[h]
  \centering
  \begin{tabular}{|p{0.3\textwidth}|p{0.65\textwidth}|}
    \hline
    \multicolumn{2}{|c|}{\textbf{Statement of Significance}} \\ \hline
    \textbf{Aspect} & \textbf{Description} \\ \hline
    \textit{Problem} & Enabling artificial agents to learn and comply with medical norms is essential for integrating autonomous systems into healthcare environments. However, the complexity of distributed healthcare systems and the dual nature of medical norms (descriptive and prescriptive) make this task particularly challenging. \\ \hline
    \textit{What is Already Known} & Existing models have been explored in domains like social norm propagation. These methods struggle to model real-world healthcare dynamics and often rely on simulated data, leading to irrelevant norms, stepwise updates, and inaccurate collective norm perception.
    \\ \hline
    \textit{What This Paper Adds} & This paper introduces the Model, which enables autonomous agents to dynamically learn descriptive and prescriptive medical norms through interaction-driven processes. Grounded in real-world data, the model prevents convergence to irrelevant norms and effectively captures descriptive norms as collective behavioral tendencies. \\ \hline
  \end{tabular}
\end{table}

\section{Medical Norm Perception and Induction}

Modeling norm perception and induction in distributed healthcare environments involves abstracting the general properties of healthcare. Establishing a general definition of descriptive and prescriptive norms within healthcare is essential. With these foundational concepts, we can naturally extend the development of multi-agent models that effectively learn these norms through robust methods.

\subsection{Descriptive Norms in Healthcare}
We standardize the expression of descriptive norms in healthcare under the term 'Medical Tendencies' (MedT).
Descriptive norms include both behavioral norms and perceptual norms. They refer to the tendency of people in their community and are less rule-like~\cite{Hawkins2019TheEO}. Behavioral norms capture the observed or communicated behavioral tendencies of a group~\cite{Eriksson2015BidirectionalAB,Miller2016ChangingNT}. Perceptual norms, in turn, refer to the communicated perceptual tendencies of a group. They describe the reference group’s perceptions or attitudes regarding behavior. We define 'tendencies' as the general characteristic of descriptive norms in healthcare.

We generalize the behavioral and cognitive tendencies of all categories of healthcare professionals (HCPs) into MedT, encompassing all scenarios, including clinical diagnosis, emergency care, nursing, and medical academic activities.

The perception of descriptive norms operates on both individual and collective levels~\cite{Legros2019MappingTS}. Individuals develop a perception of collective norms based on the information they acquire from their environment. This individual perception may be prone to errors~\cite{Miller2023ACO}. For example, a patient who survived acute coronary syndrome without undergoing percutaneous coronary intervention (PCI) may perceive PCI as ineffective or unnecessary, even though the reality might be quite the opposite. In studies on ACS treatment, clinical pathways that include timely and effective PCI represent the optimal treatment strategy and are most aligned with clinical recommendations~\cite{Kovalchuk2017SimulationOP}.
Therefore, modeling medical descriptive norms involves both individual and collective levels.

In a given healthcare environment or scenario, there exist $N$ agents, which are divided into $K$ groups. Each group has a mean Medical Tendency (MedT) : $\mu_{\text{group}}^i$ for a specific medical object. The agents within each group share the same standard deviation of MedT, denoted as $\sigma_{\text{group}}$. Based on the mean and standard deviation, a MedT value is sampled for each agent within the group. The objective collective norm (OBJ) for each group is formed by the initial objective weighted sum $w_i$ of these groups, creating an objective collective norm based on a mixture of Gaussian distributions~\cite{Falandays2021TheEO,Toscano2010CueIW}.
\begin{eqnarray}\label{eq:obj}
\text{OBJ}(x|\lambda) & = & \sum_{i=1}^{K} w_i \cdot g(x|\mu_{\text{group}}^i, \sigma_{\text{group}})
\end{eqnarray}
In this context, $\lambda = \left\{w_i, \mu_{\text{group}}^i, \sigma_{\text{group}}\right\}$ represents the set of parameters, where $i$ ranges from $1$ to $K$, corresponding to the weight, mean, and standard deviation of each group within the model, respectively.

Each agent's tendency (MedT) is modeled as a single Gaussian distribution based on the sampled MedT value from their respective group. Each agent $j$ has a unique mean $ \mu_{\text{MedT}}^j $ and a shared standard deviation $ \sigma_{\text{indiv.}} $ across all agents.
\begin{eqnarray}\label{eq:MedT}
\text{MedT}_j & \sim & \mathcal{N}(\mu_{\text{MedT}}^j, \sigma_{\text{indiv}})
\end{eqnarray}

Each agent's medical tendency, represented by its Gaussian distribution, is communicated to other agents within the healthcare scenario through medical or daily activities. This interaction is modeled as a fully connected network, where agents infer the collective norms based on the messages received from others. This subjective inference is continuously updated during interactions, rather than being a static conclusion. We define this dynamically evolving inference as Subjective Individual Norm Perception (SINP).
\begin{eqnarray}\label{eq:SINP}
\text{SINP}_j(x|\lambda) & = & \sum_{i=1}^{K} w_{\text{sub}}^i \cdot g\left(x|\mu_{\text{sub}}^i, \sigma_{\text{sub}}\right)
\end{eqnarray}
Each distribution is characterized by a subjective mean $\mu_{\text{sub}}^i$, a subjective standard deviation $\sigma_{\text{sub}}$, and a subjective weight $w_{\text{sub}}^i$, where "subjective" refers to the agent's inference about group $i$.

\subsection{Perceptions of Medical Descriptive Norms}
Individuals continuously infer collective norms through interactions, updating the subjective parameters $\lambda_{\text{SINP}}$. The convergence of an agent's SINP parameters with the OBJ parameters ($\lambda_{\text{SINP}} \rightarrow \lambda_{\text{OBJ}}$) represents the outcome of the agent's perception (learning) of the collective objective norms.

Before individuals in a group enter a healthcare environment for the first time, they naturally possess prior knowledge of norms due to social and educational influences~\cite{Morris2015NormologyII,Goldstein2007UsingSN}. An individual can obtain only limited information through each interaction within the collective. 

\begin{figure}[h]
  \centering
  \includegraphics[width=1.0\linewidth]{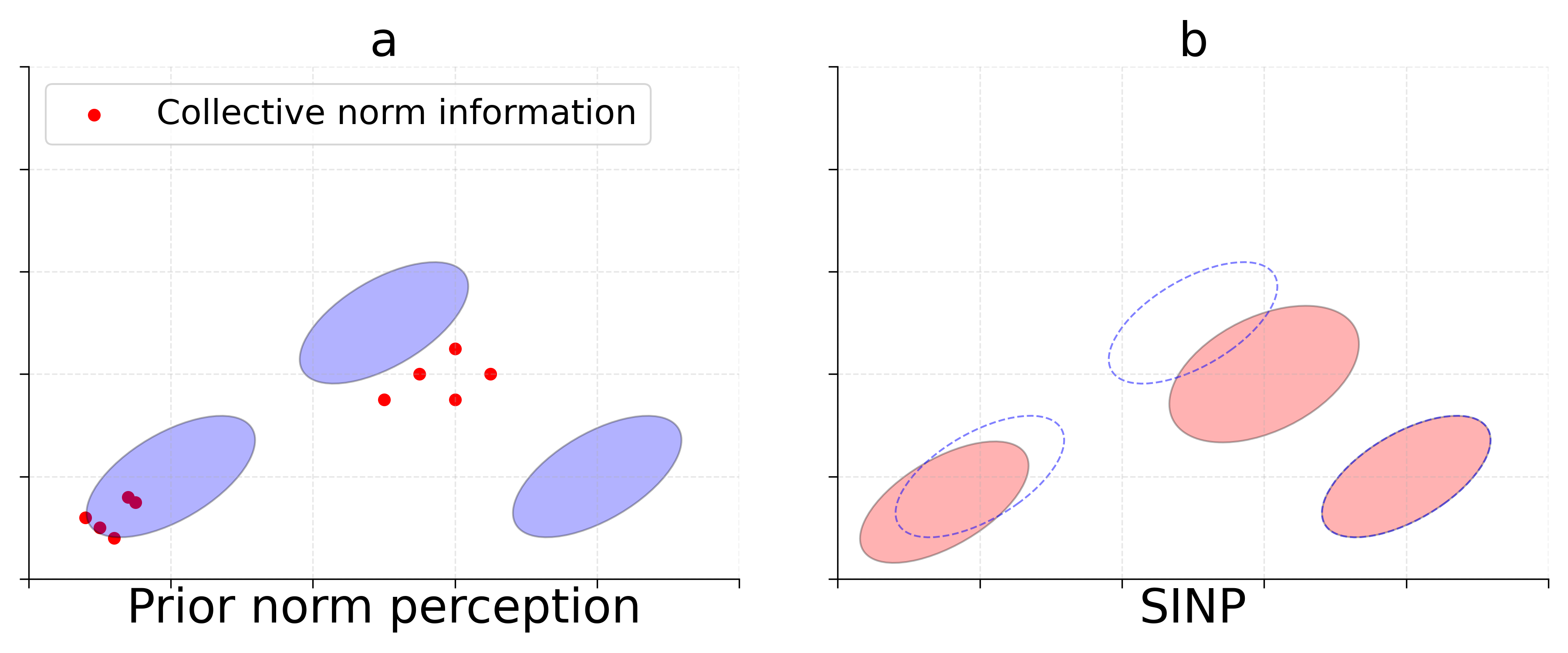}
  \caption{\textbf{The Two-Step Process of Norm Perception in Agents.} \textnormal{The model establishes SINP through two steps: (a) Collective information is mapped to prior norm perception. (b) New data is combined with prior perceptions to adapt the mixture parameters, with adjustments based on data amount.}}
  \label{fig:Perception}
  % \Description{Two-step process of norm perception: mapping to prior norm and adapting parameters.}
\end{figure}

We first derive the prior knowledge~\cite{Reynolds2000SpeakerVU} of healthcare professionals (HCPs) regarding norms based on the EHR dataset, and then parameterize this knowledge as $\lambda_{\text{SINP}}$. Next, we use Maximum Likelihood (ML) estimation to find the model parameters that maximize the likelihood of the Subjective Individual Norm Perception (SINP) given the training data, as shown in Figure~\ref{fig:Perception}. 

Our norm perception process is based on our proposed definition of descriptive medical norms, incorporating the characteristics of distributed healthcare environments, and has been adaptively modified from the standard Gaussian Mixture Model (GMM) and Expectation-Maximization (EM) algorithm.

The information that agents receive through interactions in the healthcare environment is modeled as training vectors $X = \{x_1, \dots, x_T\}$. $X$ represents a time-series array of medical tendency information received by the receiver agent from multiple sender agents, spanning from time step $1$ to time step $T$. Assuming the independence of vectors (a common assumption, albeit often incorrect, to keep the problem tractable), the likelihood of SINP can be expressed as: 
\begin{eqnarray}\label{eq:likelihood}
p(X|\lambda_{\text{SINP}}) & = & \prod_{t=1}^{T} p(\mathbf{x}_t|\lambda_{\text{SINP}})
\end{eqnarray}

 We allow agents in the healthcare environment to continually receive information about medical tendencies through norm-sharing channels and update $\lambda_{\text{SINP}}$ iteratively using the Expectation - Maximization (EM) algorithm at each time step corresponding to a medical event in the interaction process.Starting from the initial model $\lambda_{\text{SINP}}$, a new model $\hat{\lambda}$ is estimated such that the following condition holds:
\begin{eqnarray}\label{eq:likelihood_compare}
p(X|\hat{\lambda}) & \geq & p(X|\lambda_{\text{SINP}})
\end{eqnarray}
The new model $\hat{\lambda}$ becomes the starting point for the next iteration. This process is repeated until a certain convergence threshold is met. The initial model is usually obtained through some form of Vector Quantization (VQ) estimation.The process $\lambda_{\text{SINP}} \rightarrow \hat{\lambda}$ represents the learning of $\lambda_{\text{OBJ}}$ by $\lambda_{\text{SINP}}$.

In each EM iteration, the following re-estimation formulas are used, ensuring that the model likelihood monotonically increases:

- Mixture weights:
\begin{eqnarray}\label{eq:weight}
\bar{w}_i & = & \frac{1}{T} \sum_{t=1}^{T} \Pr(i|\mathbf{x}_t, \lambda)
\end{eqnarray}

- Means:
\begin{eqnarray}\label{eq:mean}
\bar{\mu}_i & = & \frac{\sum_{t=1}^{T} \Pr(i|\mathbf{x}_t, \lambda) \mathbf{x}_t}{\sum_{t=1}^{T} \Pr(i|\mathbf{x}_t, \lambda)}
\end{eqnarray}

- Variances (diagonal covariance):
\begin{eqnarray}\label{eq:variance}
\bar{\sigma}_i^2 & = & \frac{\sum_{t=1}^{T} \Pr(i|\mathbf{x}_t, \lambda) \mathbf{x}_t^2}{\sum_{t=1}^{T} \Pr(i|\mathbf{x}_t, \lambda)} - \bar{\mu}_i^2
\end{eqnarray}
Here, $\sigma_i^2$, $\mathbf{x}_t$, and $\mu_i$ represent the elements of the variance vector, the training sample vector, and the mean vector, respectively.

The posterior probability $\Pr(i|\mathbf{x}_t, \lambda)$ is given by the following equation:
\begin{eqnarray}\label{eq:posterior}
\Pr(i|\mathbf{x}_t, \lambda) & = & \frac{w_i g(\mathbf{x}_t|\mu_i, \Sigma_i)}{\sum_{k=1}^{M} w_k g(\mathbf{x}_t|\mu_k, \Sigma_k)}
\end{eqnarray}

KL divergence is used to measure the informational difference between two probability distributions, thereby directly quantifying the overlap in probability densities between SINP and OBJ. It evaluates the agent’s overall understanding of collective descriptive norms, ensuring that their behaviors and decisions align with the probability distribution of medical guidelines and best practices. This is crucial in healthcare decision-making, as healthcare professionals must ensure that their decision-making probabilities match the clinical best practices to avoid deviating from collective descriptive norms.

The KL divergence between the two distributions is computed as follows:

\begin{eqnarray}\label{eq:kl_divergence}
D_{\text{KL}}(P_{\text{SINP}} \parallel P_{\text{OBJ}}) & = & \sum_{i=1}^{K} P_{\text{SINP}}(i) \log \left( \frac{P_{\text{SINP}}(i)}{P_{\text{OBJ}}(i)} \right)
\end{eqnarray}
Where $ P_{\text{SINP}}(i) $ and $ P_{\text{OBJ}}(i) $ represent the distributions over the norm components for the subjective and objective norms, respectively. Convergence is achieved when $ D_{\text{KL}} $ falls below a predetermined threshold $ \epsilon $, signifying that $ \lambda_{\text{SINP}} $ has successfully learned $ \lambda_{\text{OBJ}} $.

\subsection{Prescriptive Norms in Healthcare}
In the context of prescriptive norms, they typically involve fundamental principles~\cite{Scharding2023WhenAN}. This is also true for medical prescriptive norms, which consist of medical rules, protocols, or guidelines~\cite{Kovalchuk2017SimulationOP}. Medical rules exist in the minds of individuals as psychological representations, akin to "concepts," and they also reflect the structured nature of the medical settings~\cite{Armstrong1983WhatSC,Smith1956ASO}.

In the study of concept learning, rational inductive logic models based on rule-based frameworks are well-suited to capturing compositional representations~\cite{Bach1998ConceptsWC,Murphy2002TheBB,Nosofsky1994RuleplusexceptionMO}. This Rational Rules model combines the inferential power of Bayesian induction with the representational power of mathematical logic and generative grammar; the former accounts for how concepts are learned under uncertainty, whereas the latter provides a compositional hypothesis space of candidate concepts~\cite{Goodman2008ARA}. Their approach offers a grammar that compactly defines an infinite and flexible hypothesis space of structured rules, starting from simple components, with a prior that effectively controls complexity.

Researches~\cite{Muggleton1994BayesianIL,Goodman2008ARA} indicate that mathematical logic provides a simple and intuitive framework for concepts. By considering rules as functions from objects to truth values (where True is arbitrarily assigned to "in the concept"), a natural set of combination operations based on logical connectives emerges. This suggests that using first-order predicate logic as fundamental components, and flexibly combining them to represent rule-based clinical prescriptive norms of varying complexity, is effective.
We consider medical prescriptive norms as functions from medical objects or states to truth values, indicating whether these objects comply with the norms.

Let $\mathcal{N} : \mathcal{O} \rightarrow {0, 1}$, where $\mathcal{O}$ is the set of medical objects. $\mathcal{N}(\mathbf{o}) = 1$ indicates that the object $\mathbf{o} \in \mathcal{O}$ meets the normative criteria, while $\mathcal{N}(\mathbf{o}) = 0$ indicates noncompliance.
\begin{eqnarray}
\mathcal{N}(\mathbf{o}) = 
\begin{cases} 
1, & \mathbf{o} \text{ meets criteria}, \\
0, & \text{otherwise}.
\end{cases}
\end{eqnarray}
Here, $\mathbf{o} = (h, s, a, s')$, which represents an agent's state-action history, where $h$ denotes the history of past states and actions, $s$ is the current state, $a$ is the action taken, and $s'$ is the resulting state. The use of first-order predicate logic to represent norms is to enable precise evaluations of an agent’s state-action history. Representing norms as functions $N \to {0, 1}$ facilitates the quantification of the state-action history set for agent $i$, defined as $H_{ist_i} := \bigcup_{t=1}^{Hist} ((S \times A_i)^t \times S)$. Similarly, $H_{ist} := \bigcup_{i \in I} H_{ist_i}$ represents the union of all agents’ history sets. By quantifying these state-action history sets, agents can infer and update their beliefs about norms based on observations of other agents’ behaviors. 

We can use first-order predicate logic to extend complex rules. In rule-based rational inductive logic models, the "hypothesis space" is a space containing all possible rules used to form or induce concepts. This space is determined by the syntactic structure of the concepts, which provides the rules for concept generation and development. This hypothesis space handles the constructive relations between more complex hypotheses and simpler ones~\cite{Goodman2008ARA}.

Let $\mathcal{H}$ represent the hypothesis space of prescriptive norms. This space is defined as the set of all possible functions $N : \mathcal{O} \rightarrow {0, 1}$. Each function $f$ in $\mathcal{H}$ maps an object $o \in \mathcal{O}$ to a truth value, indicating compliance with a norm.
\begin{eqnarray}
\mathcal{H} = \left\{ N : \mathcal{O} \rightarrow \{0, 1\} \right\}
\end{eqnarray}
Using first-order language $L$ rules and incorporating the advice of medical experts, we enumerated all possible predicates and numerical thresholds to generate the norm space $H$.  inferring precise posterior probabilities over $H$ is highly intractable, as it requires enumerating all subsets $H' \subseteq H$ and updating their probabilities. This computational complexity cannot be resolved by simply controlling the generation of $H$. Therefore, our work employs a mean-field approximation to address this issue. 

\subsection{Learning Medical Prescriptive Norms}
In the field of healthcare modeling, discrete-event simulation~\cite{Klein2012EmergencyDP,Yang2013AMT}, system dynamics~\cite{Lane2003ClientII,Tian2016LinkedSA}, agent-based modeling~\cite{Isern2016ASL}, and queueing theory~\cite{Cochran2006StochasticBB,deBruin2005BottleneckAO} are popular approaches, each addressing different types of distributed problems in medical environments.

Based on the various complex distributed problems in healthcare scenarios that are described and addressed by these popular approaches, we have found that enabling autonomous agents to continuously interact within medical environments, and using rational inductive logic models to derive prescriptive norms from sparse and noisy data in uncertain conditions, greatly facilitates the ability of autonomous agent systems to collaborate effectively with HCPs.

Combining Markov games with a model of "approximate" rational Bayesian structured rule induction~\cite{Tan2018ThatsML,Cranefield2015ABA,Nichols2016RationalLA} has shown inaccuracies in general social norm learning experiments~\cite{Oldenburg2024LearningAS}, where agent systems may learn nonfunctional or even irrational prescriptive norms. 

Based on the results observed in our rule (prescriptive norms) induction experiments on the medical dataset, we hypothesize that this phenomenon may be due to two factors. First, the probabilistic nature can lead to the emergence of low-frequency rules. While this is possible in the model, it is usually rare in real life (typically due to external regulation rather than the internal nature of the system). Second, self-supporting groups of agents may emerge within complex systems, which internally increase the frequency of certain rules. This is similar to the information bubbles and echo chambers studied in social networks~\cite{Bazhanov2022CognitiveMI,Kitchens2020UnderstandingEC,Interian2022NetworkPF}. Moreover, this prevents the system from converging to a single behavior pattern represented by a unified normative space (the system converges to multiple variants).

Therefore, considering the seriousness of the healthcare context, we introduced external regulation in the form of "practice verification" into the rational inductive logic model based on Markov games. Additionally, we incorporated adaptive learning rates and momentum mechanisms into the process of agents' concept updates on prescriptive norms, which are based on Bayesian inference~\cite{Duchi2011AdaptiveSM, Rumelhart1986LearningRB,Kingma2014AdamAM}.

In the context of the prescriptive norm space $\mathcal{H}$, we extend the general setting of multi-agent Markov games by formulating judgments about the prescriptive norm $N(o)$ based on the Bellman Equation. Here, $s \in S$ represents the state space, $a_i \in A_i$ denotes the action taken by agent $i$, and $T_i(s'|s, a_i)$ is the transition function that defines the probability of moving from state $s$ to state $s'$ given action $a_i$. The reward function $R_i^{N}(s, a_i, s')$ captures the immediate reward for agent $i$ under the prescriptive norm $N$, and $\gamma$ is the discount factor.

The action-value function $Q_i^{N}(s, a_i)$ and the state-value function $V_i^{N}(s)$ are then defined as follows:
\begin{eqnarray}\label{eq:Q_function}
Q_i^{N}(s, a_i) & = & \sum_{s' \in S} T_i(s'|s, a_i) \left[ R_i^{N}(s, a_i, s') + \gamma V_i^{N}(s') \right]
\end{eqnarray}
\begin{eqnarray}\label{eq:V_function}
V_i^{N}(s) & = & \max_{a_i \in A_i} Q_i^{N}(s, a_i)
\end{eqnarray}

Our model is based on real-world EHR datasets provided by healthcare organizations and incorporates the guidance of medical experts who created these datasets. Consequently, the norms within our prescriptive norm space $\mathcal{H}$ are derived from authentic medical protocols.

The total reward function for agent $i$ when taking action $a_i$ in state $s$ and transitioning to state $s'$ is represented as $R_i^{\mathcal{H}}(s, a_i, s')$, which is defined as follows:

\begin{align}\label{eq:reward_function}
R_i^{\mathcal{H}}(s, a_i, s') = C_{a_i}^{\text{medical}} + \sum_{n_m \in \mathcal{H}} \left[C_i(n_m) + r_i(n_m)\right] N_m(s, a_i, s')
\end{align}
Here, $R_i^{\mathcal{H}}(s, a_i, s')$ considers both the medical cost associated with the action, $C_{a_i}^{\text{medical}}$, and the cumulative effect of complying with norms within the prescriptive norm space $\mathcal{H}$. The summation term captures the combined influence of the compliance cost $C_i(n_m)$ and the reward $r_i(n_m)$ for fulfilling the norm $N_m$. The indicator function $N_m(O)$, where $O=(s, a_i, s')$, determines whether this outcome meets the prescriptive norm criteria.Here, $m$ is an abbreviation for "medical" and does not have any other meaning.

The induction of rules (prescriptive norms) is reflected in the update of agents' beliefs about these norms. Based on prior related research~\cite{Oldenburg2024LearningAS,Goodman2008ARA,Tan2018ThatsML}, our calculation method extends three assumptions: 1. Agents are aware of the existence of the entire norm space $\mathcal{H}$ but hold uniformly low prior beliefs about each norm.
2. Norms are shared, and agents model each other as adhering to these norms.
3. Agents assume that other agents will act deterministically when following a specific norm $n_m$.

This update is derived from rational Bayesian inference, summarizing the agents' beliefs about the norms from a large set of discrete data points in their $Q$ and $V$ tables. We calculate the local posterior probability for each $N \in \mathcal{H}$ using a mean-field approximation. The likelihood term $\pi_{-i}(a_{-i} | h, s, n_m)$ represents the probability that other agents take their most recent actions \(a_{-i}\) given that $n_m$ is the only known norm at the current time step and is being followed by agent $a_i$. We assume that each agent $j$ selects its action $a_j$ according to a Boltzmann distribution. 
\begin{eqnarray}\label{eq:P_update}
\tilde{P}_i^{t+1}(n_m | h, s, a_{-i}) &\propto& \pi_{-i}(a_{-i} | h, s, n_m) \tilde{P}_i^t(n_m) \\
\pi_{-i}(a_j | h, s, n_m) &\propto& \exp(Q^{n_m}(s, a_j))
\end{eqnarray}
Here, $Induction$ refers to the process by which agents in our system infer norms through approximately rational Bayesian rule induction based on observations of compliance and violations of potential norms. As described in Assumptions 2 and 3, as well as Equations (16) and (17), agents update the probabilities of each potential norm by observing the behaviors of other agents. Each agent assumes that others may follow shared norms while pursuing their own interests, inferring norms from seemingly non-beneficial behaviors and aggregating observations to update their beliefs about the norms that best explain collective behavior.

Assumption 1 ensures that the norm verification and adjustment process is integrated into the agents' belief updating mechanism, where beliefs $b(n_m)$ about norms $n_m \in \mathcal{H}$ are adjusted by a regulation factor $\alpha$ based on observed efficacy and normalized to form a valid probability distribution, as shown below:
\begin{eqnarray}\label{eq:belief_normalization} b_i''(n_m) & = & \frac{\alpha \cdot b_i(n_m)}{\sum_{n_m \in \mathcal{H}} (\alpha \cdot b_i(n_m))} \end{eqnarray} 
Detailed explanations for this process are provided in Appendix~\ref{eq:belief_update} and Appendix~\ref{eq:belief_normalization}. Furthermore, adaptive learning rates $\alpha_{i}^{t+1}(n_m)$ (Appendix~\ref{eq:momentum_update}) and momentum $m_{i}^{t+1}(n_m)$ (Appendix~\ref{eq:adaptive_lr}) are incorporated to stabilize updates and prevent convergence to local optima, with the learning rate dynamically adjusted using second moment estimates $v_{i}^{t+1}(n_m)$ (Appendix~\ref{eq:second_moment}). Detailed formulas and explanations are provided in the Appendix~\ref{appendix:Norm Practice and Smooth Updates}.

\section{EXPERIMENTS}
We reviewed the general internal structure of both the medical norm system and the norm-sharing channels system in the healthcare domain. The implementation of specific norm categories within the medical norm system is achieved through simulation and experimentation in combination with the dataset. These two systems are mapped onto our formal model through specific examples.

\subsection{Medical Norm Systems and Their Sharing Channels}
The system of medical norms includes protocols, clinical recommendations, medical standards, etc. (Table~\ref{tab:categories}). Norm sharing channels, including official requirements, best practices, guidelines, etc., disseminate medical standards to healthcare professionals (Table ~\ref{tab:sharing_channels}). 
Tables ~\ref{tab:categories} and ~\ref{tab:sharing_channels} outline the main categories that constitute the two systems. These are not intended to cover all categories exhaustively but to explain the general and primary components.

Protocols focus on standardized procedures~\cite{Ralli2020HashimotosTA,Chung2020RevisedTA}, while clinical recommendations offer flexible suggestions based on current research and evidence~\cite{Kasper2020PracticalRF,Phi2012ExpandingTG}. Medical standards are benchmarks set by governments or organizations to ensure medical service quality and safety, carrying legal enforceability~\cite{Chandran2021DevelopmentOT}.

In the norm sharing system, official requirements are enforced through laws and policies, ensuring uniform standards. Best practices are shared via conferences, journals, and associations, often proposed by experts. Guidelines, established by professional organizations (such as the American Heart Association or the World Health Organization), are distributed globally through official publications, education, and online databases, providing physicians with systematic, evidence-based decision-making tools~\cite{Holbrook2012EvidencebasedMO,CruzRivera2022EthicalCF,Agnese2019NCCNCP}.

\FloatBarrier
\begin{table}[H]
  \caption{The system of medical norms}
  \label{tab:categories}
  \begin{tabular}{p{0.25\linewidth}p{0.65\linewidth}}\toprule
    \textbf{Category} & \textbf{Sub-categories with References} \\ \midrule
    \textit{Protocols (P)} & Diagnostic P~\cite{Ralli2020HashimotosTA,Carruthers2003MyalgicEC,Fratz2013GuidelinesAP,Than2014DevelopmentAV,Bobdiwala2018DiagnosticPF}, Therapeutic P~\cite{Cotovio2023RegulatoryCA,Dar2014StudiesOE,Ghaffari2010OptimalTP}, Emergency P~\cite{Chung2020RevisedTA,Long2020ModificationsOE,Hall2020DoCA}, Preventive Care P~\cite{Litzelman1993RequiringPT,Frydrych2017ComplianceOP,Banerjee2016OptimizingPM}, Rehabilitation P~\cite{Li2023ReviewOR,DvilaCastrodad2019RehabilitationPF,Zanini2019EffectsOD,Rucinski2019EffectsOC} \\ \midrule
    \textit{Medical Recommendations (MR)} & Personalized MR~\cite{Williams2022PersonalizedT2,Wischmeyer2023PersonalizedNT,Kamath2022ImprovingOT,Hu2023PersonalizedHT,Chatterjee2021PersonalizedRF,Zhou2019MultiModalityBI}, Ethical and Safety MR~\cite{Rossi2020SafetyAR,Scendoni2023TheSO,Bell2020ScientificAE,Prigoff2016EthicalRF}, Population-specific Patient MR~\cite{Kasper2020PracticalRF,Mihalj2020RecommendationsFP,Hert2011PhysicalII}, Evidence-Based Grading MR~\cite{Phi2012ExpandingTG,Interian2022NetworkPF,Moran2020EvidenceSA}, Multidisciplinary Collaboration MR~\cite{Andersen2017APF,Ligibel2017AmericanSO,Santiago2022RecommendationsFT} \\ \midrule
    \textit{Medical Standards (MS)} & Care and Operations MS~\cite{Chandran2021DevelopmentOT,Basevi2010StandardsOM,Heeren2023ACS,altevogt2009operational,Mills2018DevelopmentAI}, Health and Safety MS~\cite{Laverty2017EmbeddingCS,Alkon2015HealthAS,Doronina2016NewIS}, Data and Information Management MS~\cite{Garcia2020CentersFD,Fenton2017HealthIM} \\ \bottomrule
  \end{tabular}
\end{table}

\begin{table}[t]
  \caption{The norm sharing channels system}
  \label{tab:sharing_channels}
  \begin{tabular}{p{0.25\linewidth}p{0.65\linewidth}}\toprule
    \textbf{Category} & \textbf{Sub-categories with References} \\ \midrule
    \textit{Guidelines (G)} & Clinical Practice G~\cite{Poser1983NewDC,Agnese2019NCCNCP,Khwaja2012KDIGOCP,Galle2018EASLCP,Kearon2012AntithromboticTF}, Treatment G~\cite{Voors20162016EG,Workowski2015SexuallyTD,Hert2011PhysicalII,Association2020JapaneseGC}, Public Health G~\cite{Moran2021PredictorsOA,Dempsey2020GlobalPH,Knight2015PublicHG}, Preventive Care G~\cite{Matthys2014PersonalizedEO,Hajizadeh2013PersonalizedEO,Overhage1996ComputerRT}, Ethics G~\cite{CruzRivera2022EthicalCF,Joebges2020EthicsGO,Mentzelopoulos2021EuropeanRC} \\ \midrule
    \textit{Best Practices (BP)} & Evidence-Based Clinical BP~\cite{Holbrook2012EvidencebasedMO,Norcross2011EvidencebasedTR,Singh2016HandbookOE,Hoagwood2014ScalingUE,McGovern2004ASO}, Patient Safety and Risk Management BP~\cite{Fusari2021BestLP,Palaian2020PatientSC,Ricciardi2020GuidelinesAS}, Quality Improvement BP~\cite{Lee2020OutcomesAC,Cohen2018PrimaryCP,Moon2017TheEO,Leff2015HomeBasedPC} \\ \midrule
    \textit{Official Requirements} & Laws and Regulations~\cite{Bierman2018StateOA,Xuejun2012BriefIO}, Health Policies~\cite{Bargain2020TrustAC,Jiang2021ChangesIO,Bhalerao2019PublicHP}, Institutional Policies~\cite{Nates2016ICUAD,Ogbuabor2018ImplementationOF,Dzeng2015InfluenceOI}, Medical Devices and Pharmaceuticals Approval~\cite{Marshall2021TheDB,Braillon2010MedicalDA,Sweet2011ReviewOT} \\ \bottomrule
  \end{tabular}
\end{table}

\subsection{Experimental Dataset}
Our experimental datasets include a dataset from a 2016-2020 vertigo clinic in Rostov-on-Don \cite{Petruchik2021}. In the dataset provided by the vertigo clinic, descriptive norms are proposed in the form of clinical recommendations advocated by experts, while prescriptive norms are enforced through internal clinic protocols. Both are shared as best practices through collaborative exchanges within the medical community

\subsection{Descriptive Norms in Vertigo Dataset}
Vertigo, particularly benign paroxysmal positional vertigo (BPPV), is a complex condition characterized by a multitude of etiologies and the involvement of various medical specialists, including neurologists and otolaryngologists, among others \cite{Petruchik2021}.

Based on our previous experiments \cite{Li2024ModellingOP} with this dataset and in consultation with the medical experts who created it, we introduced the \textit{Practice Shannon Diversity Index}, calculated from practice preference vectors \cite{Li2024ModellingOP}, to represent the openness of 10 neurologists in the dataset toward various vertigo diagnostic approaches. This openness is incorporated into the framework of descriptive medical norms and modeled as the Medical Tendency (MedT) of agents within this norm system.

We aggregated and processed all diagnostic information spanning 5 years from the dataset. Subsequently, we applied K-means clustering to group the overall Practice Shannon Diversity Index for the 10 neurologists across these 5 years. The results, as depicted in the corresponding elbow method plot, indicate that the $WCSS$ (within-cluster sum of squares) levels off after $K=5$ (See details in Appendix~\ref{fig:combined_clusters_elbow}). To avoid overfitting and excessive complexity in the grouping, we selected $K=5$.

We calculated the average Practice Shannon Diversity Index and variance for each of the 5 clusters, and initialized the collective objective descriptive norms for the vertigo treatment scenario based on the proportion of doctors in each cluster. These norms were modeled as a Gaussian Mixture Model. 

Next, we simulated the interactions of $n$ agents within the same scenario. This simulation modeled the process by which the clinic’s doctors shared their treatment tendencies for vertigo through medical practice sharing activities, incorporating best practices within the medical community. We modeled all agents as a fully connected network. At each time step, half of the agents are randomly selected as receivers, while the other half act as senders. The receivers can acquire information about the Practice Shannon Diversity Index generated by the senders within the network during the current time step (i.e., the medical practice sharing activity). Based on the information received, the receiver agents adjust their Subjective Individual Norm Perception (SINP) dynamically. The data $X$ received by agent $j$ at time $t$ is used to compute its Gaussian Mixture Model (GMM, $SINP_t$), where the subjective mean, standard deviation, and weight are iteratively inferred using the EM algorithm as described in Equations 4–9. Agent $j$ maximizes the likelihood of the data to iteratively update its subjective perception of $OBJ$, including the subjective mean, standard deviation, and weight, gradually aligning with the distribution of the received data. By accumulating sufficient $X$ data over time, the agent optimizes its GMM ($SINP_j$), progressively approximating the objective norm $OBJ$. 

\begin{figure}[h]
  \centering
  \includegraphics[width=0.75\linewidth]{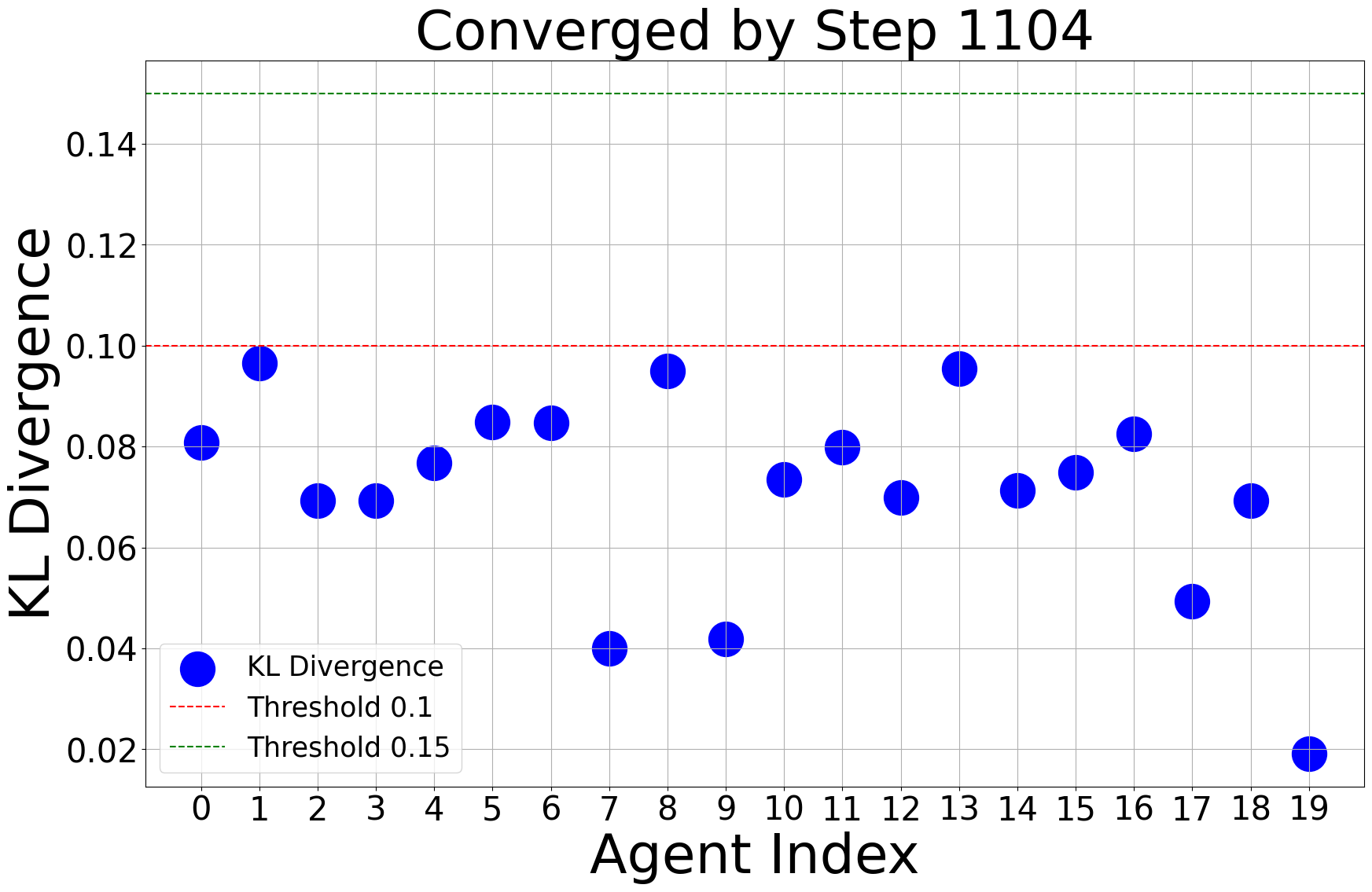}
  \caption{KL Divergence between $\text{SINP}_i$ and OBJ after practice sharing activities.}
  \label{fig:kl_divergence_medical_practice_sharing_activities}
  % \Description{(a) KL Divergence values over time steps, where the curve drops below 0.1 after 1104 steps, indicating convergence. (b) A visualization showing SINP convergence in a multi-agent system.}
\end{figure}

In Figure~\ref{fig:kl_divergence_medical_practice_sharing_activities}, we present the results from a simulation involving 20 agents. After 1104 time steps (representing practice-sharing activities), the KL divergence between an agent’s $\text{SINP}_i$  and the OBJ across the community fell below 0.1. This indicates that the medical agents ensured their decision-making probabilities were aligned with the collective medical norms representing the current best clinical practices in the environment. 

\begin{figure}[h]
  \centering
  \includegraphics[width=0.75\linewidth]{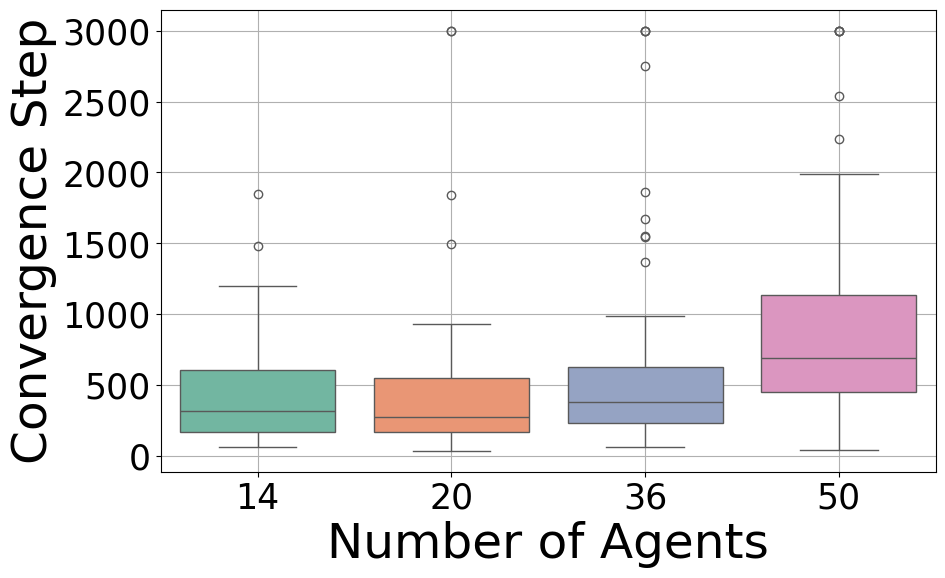}
  \caption{Boxplot showing the distribution of convergence steps for different numbers of agents.}
  \label{fig:boxplot_convergence_steps}
  % \Description{A boxplot visualizing the convergence steps for agent counts of 14, 20, 36, and 50, illustrating the variation in convergence times.}
\end{figure}

\begin{figure}[h]
  \centering
  \includegraphics[width=0.75\linewidth]{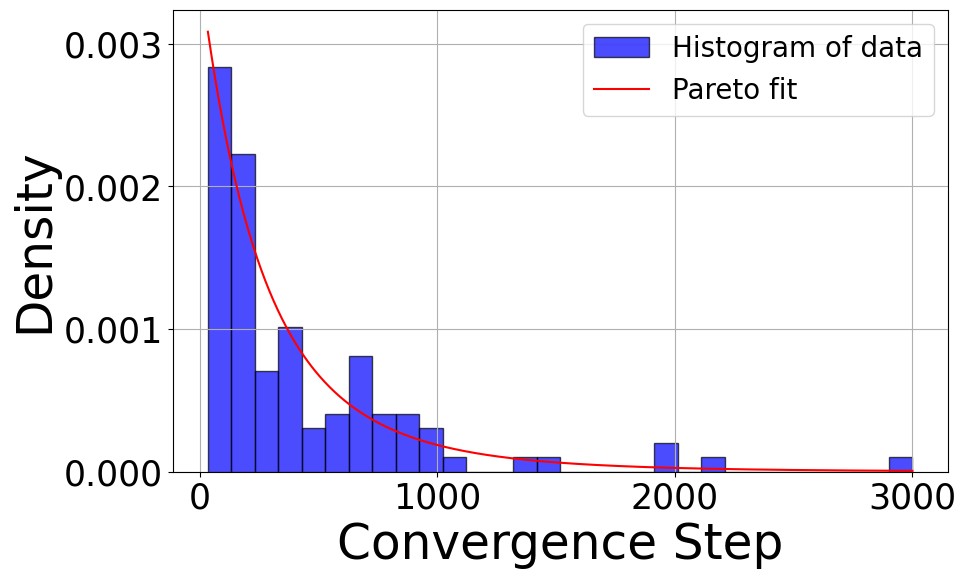}
  \caption{Histogram of Convergence Steps for 20 Agents over 50 runs.}
  \label{fig:ConvergenceSteps_20Agents_50run}
  % \Description{Histogram showing the distribution of convergence steps across 20 agents during 50 simulation runs, with most agents converging within 3000 steps.}
\end{figure}

We conducted 50 independent experiments for each case with 14, 20, 36 and 50 agents (Figure ~\ref{fig:boxplot_convergence_steps}). The distribution fitting and Kolmogorov-Smirnov test were performed on all results. The convergence time distribution for 14 agents closely followed the lognormal distribution, while for 20 agents, it aligned with the Pareto distribution (Figure ~\ref{fig:ConvergenceSteps_20Agents_50run}). The distributions for 36 and 50 agents were closer to the Burr distribution. Notably, all these distributions fall under the category of heavy-tailed distributions (Figure details are in Appendix~\ref{fig:lognorm_14_agents},~\ref{fig:burr_36_agents},~\ref{fig:burr_50_agents}).

\begin{figure}[h]
  \centering
  \includegraphics[width=0.75\linewidth]{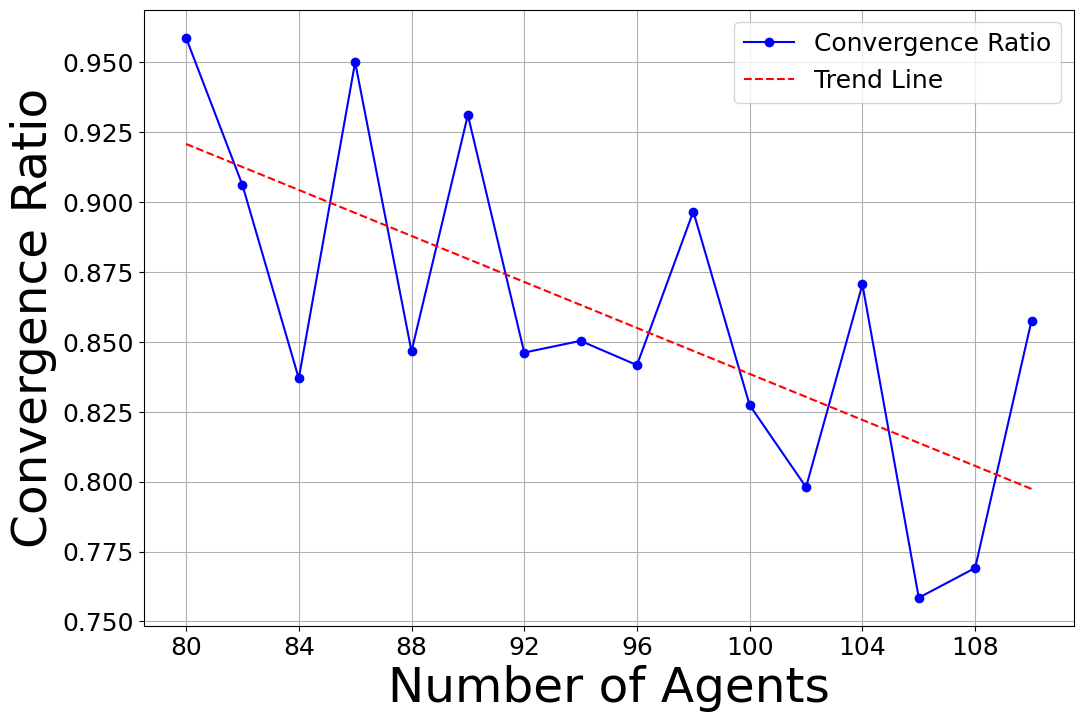}
  \caption{Convergence Ratio Decrease with Increasing Number of Agents.}
  \label{fig:convergence_ratio_agents}
  % \Description{This figure shows the relationship between the number of agents and the convergence ratio. As the number of agents increases, the system's ability to reach full convergence decreases, reflecting the challenges of achieving a 100 percent convergence rate with a larger number of agents.}
\end{figure}

We incrementally increased the number of agents in the system from 80 to 110, with a step size of 2, and conducted 10 repeated simulations for each agent configuration. The results, as shown in Figure~\ref{fig:convergence_ratio_agents}, present the average convergence ratio for each set of simulations. Although there are slight fluctuations in certain intervals (e.g., between 80 and 92 agents), the overall trend clearly demonstrates that as the number of agents increases, the convergence ratio decreases, dropping from nearly 0.95 to approximately 0.75. Even in a fully connected network, some agents failed to converge. 

This can be attributed primarily to the randomness in information exchange at each time step. Particularly in cases with a large number of agents, individual agents may not receive complete information from all other agents, leading to insufficient or incomplete data over extended periods, which delays convergence. Moreover, at each time step, only a subset of senders can be sampled, resulting in potential delays in information transmission, especially when the agent count exceeds 80. This delay affects the agents' update process. Specifically, agents whose initial SINP parameters differ significantly from the OBJ parameters, and those that were initialized with suboptimal cluster centers, tend to have slower convergence rates or may fail to converge altogether.

The Practice Shannon Diversity Index (SDI) of 10 doctors in the dataset, after undergoing 17 practice sharing activities over 5 years, shows a trend where the SDI values of chief physicians and assistant physicians gradually converge. These phenomena highlight how descriptive norms (practice tendencies) in medical scenarios gradually converge through dynamic interactions, as shown in the model demonstrated in this Section. In real-world healthcare, long-term practice-sharing activities foster greater consistency in practices and decision-making, reflecting the gradual formation of collective norms as doctors adopt similar diagnostic and treatment strategies.

In the dataset, the average SDI increment among 10 doctors steadily rose from 2016 to 2018, likely due to new sharing activities and practice schemes that broadened perspectives and techniques. From 2018 to 2020, the increment declined, suggesting stabilization of collective norms and reduced exploration of new practices, with descriptive norms reaching a consistent and stable state within the community(see details in Appendix~\ref{fig:shannon_diversity_all_doctors},~\ref{fig:average_shannon_increment}). 

\begin{figure}[h]
  \centering
  \includegraphics[width=0.75\linewidth]{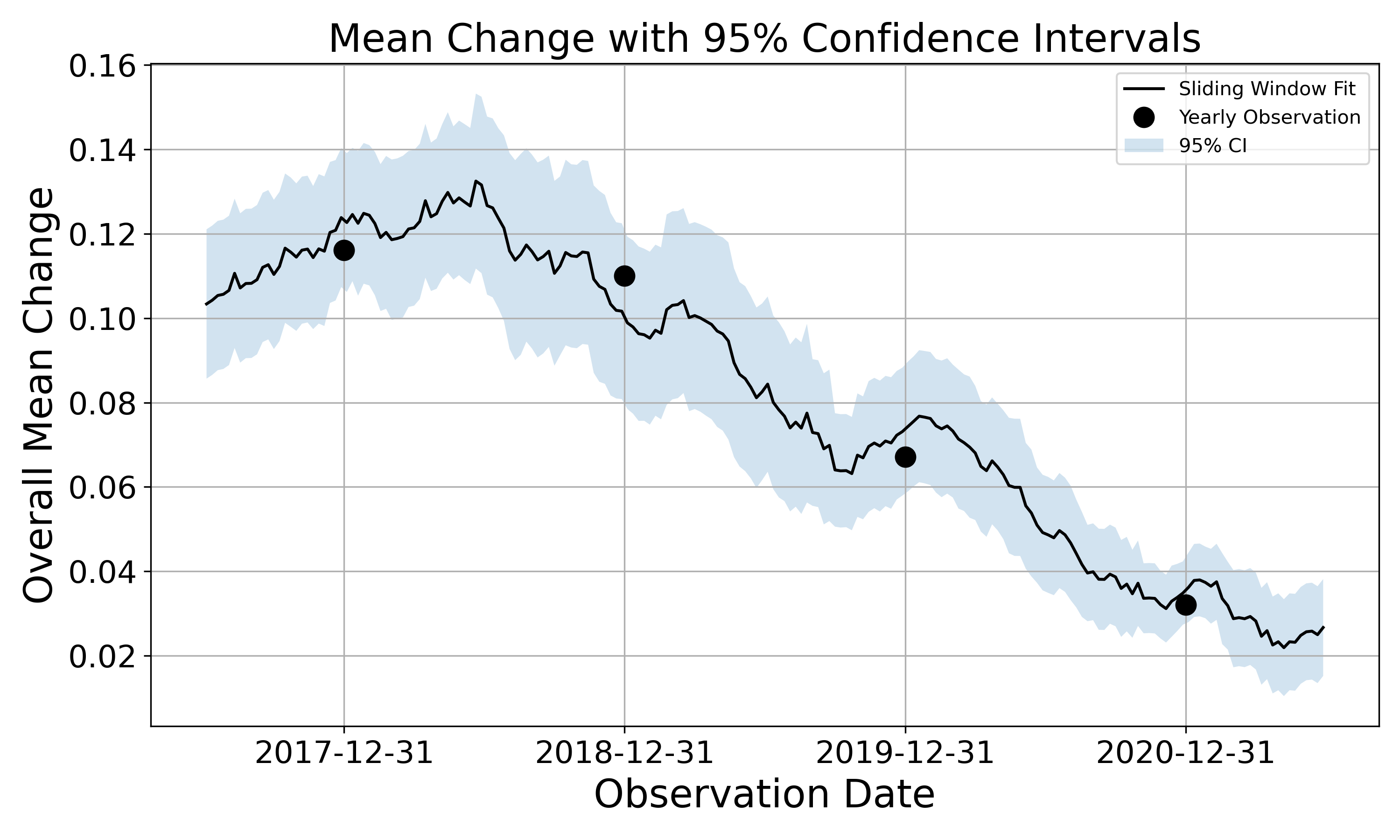}
  \caption{Mean Change over Time with 95\% Confidence Intervals and Yearly Observations (black dots).}
  \label{fig:mean_change_sliding_window}
  % \Description{This plot shows the mean change over time, smoothed using a sliding window. It includes 95\% confidence intervals and yearly observations, represented by black dots.}
\end{figure}

We modeled this process by simulating the SDI changes for each two-year period. The system’s SINP was initialized using the SDI data of the respective year, and the system then learned the OBJ for the following year. Each two-year experiment was repeated 50 times. A sliding window method was applied to smooth the average SDI increment data for each agent in the experiments, producing the trend curve shown in Figure~\ref{fig:mean_change_sliding_window}. Observation points summarize the year’s data collected on the last day of each year. 

In the experiment, each cycle spanned two years, where SINP was initialized using the SDI of the respective year and then learned the OBJ for the following year. This staged simulation effectively captured both short-term dynamics and long-term trends in the formation of descriptive norms. A sliding window method was applied to smooth the SDI increment data, generating clearer trend curves by filtering out short-term fluctuations and highlighting the long-term changes in descriptive norm formation.

Data collected on the last day of each year were summarized as observation points to illustrate annual changes. This design aligned experimental data with specific time points, facilitating the analysis of the temporal pathways in the formation of descriptive norms.

Overall, the dataset and the results shown in Figure~\ref{fig:mean_change_sliding_window} emphasize the significance of dynamic interactions. Practice-sharing activities, as the driving force for the convergence of descriptive norms, enable the medical community to identify and adopt better practices by sharing the experiences of different doctors. This dynamic interaction mirrors the real-world process of norm formation in medical settings, driven by collaboration, communication, and knowledge sharing.

\subsection{Prescriptive Norms in Vertigo Dataset}
Based on the professional recommendations from the dataset contributors, we identified that the most common clinical practice in the vertigo outpatient department combines Romberg, Hallpike, and StepTest for diagnosis. These three tests assess different systems (vestibular, nervous, and motor balance) in vertigo patients, allowing physicians to distinguish between types of vertigo and accurately diagnose the cause, location, and nature of the condition. Accordingly, we identified two key protocols within the clinic and their corresponding predicate logic representations (see Table~\ref{tab:protocols}).

In addition to the clinical protocols represented by norms $\mathcal{N}_1$ and $\mathcal{N}_2$, we introduced 30 control norms generated from combinations of physician agent types, diagnostic procedures, and time steps. In each experiment, 6 control norms are randomly selected along with the two core protocols for the multi-agent system to learn. Examples of the Hypothesis Space of Prescriptive Norms and the complete set of control norms can be found in Appendix~\ref{fig:Hypothesis_Space_H} and Appendix~\ref{appendix:control norms}.

We extracted the top 7 most commonly used diagnostic methods from the dataset. We define the exploration space for the multi-agent system as a Markov game environment, consisting of combinations of randomly selected diagnostic methods, physician agent roles, and time steps.

We configure all agents to execute Real-Time Dynamic Programming (RTDP) with a depth of 3 every 2 time steps for exploration. Every 15 time steps, agents sample new norms from the norm space and adhere to them. Each time an agent uses a diagnostic method, it incurs a diagnostic cost of 0.1. Agents incur a prescriptive norm violation cost of 3 for failing to comply with protocol $\mathcal{N}_1$ and 1 for failing to comply with protocol $\mathcal{N}_2$. Similarly, agents receive a reward of 1 for complying with either protocol $\mathcal{N}_1$ or $\mathcal{N}_2$. Additionally, the more diagnostic methods an agent uses, there is a 50 percent chance of receiving an extra reward of 0.5 from patient agents. Agents select actions based on a Boltzmann distribution.

Agents need to update their beliefs about a specific prescriptive norm based on the observed behavior of other agents in the healthcare environment. This is a learning process based on Bayesian inference, where agents adjust their belief values for the norm after each observation. In other words, agents infer rules from the data generated by the multi-agent system in the Markovian environment and further practice and verify the learned prescriptive norms within the environment. The adjustment factor $\alpha$ is responsible for reweighting and scaling the belief $b_i(n_m)$ that agents hold about each norm $n_m \in \mathcal{H}$ based on their practical validation of the norm. To make this adjustment process efficient and stable, we introduced adaptive learning rate adjustments. By calculating both the first-order momentum and second-order moment estimates (variance), the learning rate is adaptively adjusted at each update step, preventing divergence due to an overly large learning rate or slow convergence due to an overly small one.

\begin{figure}[h]
  \centering
  \includegraphics[width=0.75\linewidth]{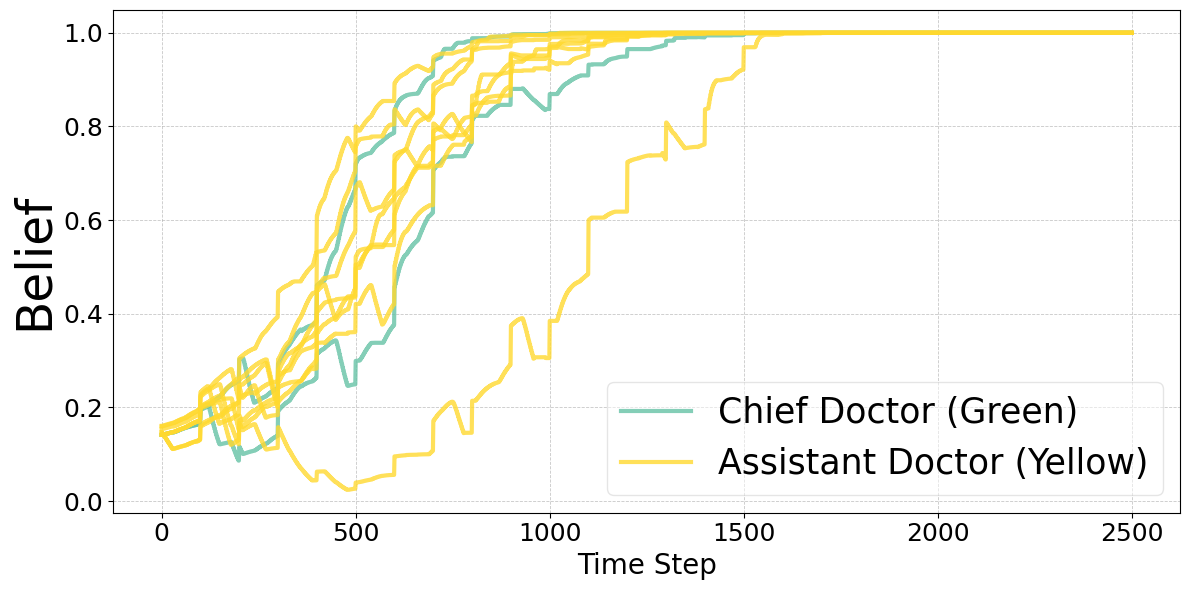}
  \caption{Belief Dynamics of Chief and Assistant Doctor Agents Regarding Protocol $\mathcal{N}_1$ over 2500 Time Steps}
  \label{fig:Belief_Dynamics}
  % \Description{A plot showing belief updates over time for chief doctor and assistant doctor agents during a 2500 time step simulation. The x-axis represents time steps, and the y-axis shows belief values, with separate lines for each agent role.}
\end{figure}

We set the discount factor $\gamma = 0.9$, the Boltzmann temperature $t = 5$, the momentum coefficient $\beta_1 = 0.9$, and the RMSprop coefficient $\beta_2 = 0.999$. The initial learning rate is set to $\alpha = 0.001$, and the practice adjustment factor is $\alpha_{\text{practice}} = 0.7$. Agents practice their prescriptive medical norms based on their beliefs every 100 time steps.Figure~\ref{fig:Belief_Dynamics} presents the belief updates for protocol $\mathcal{N}_1$ over 2500 time steps in a multi-agent system consisting of 10 agents, including 2 chief doctor agents and 8 assistant doctor agents. As shown in Figure~\ref{fig:Belief_Dynamics}, we significantly reduced stepwise updates in the belief dynamics for norms. Additionally, agents with different roles in the system successfully learned the core norms within approximately 1500 time steps.

\begin{table}[t]
    \caption{Core Protocols and Their Predicate Logic Representations}
    \label{tab:protocols}
    \begin{tabular}{p{0.9\linewidth}}\toprule
        \textbf{N1}: Doctors must use the top 3 diagnostic tests. \\
        $\forall a (Role(a, Doctor) \implies Uses(a, \{Romberg, Hallpike, StepTest\}))$ \\ \midrule
        \textbf{N2}: Chief and assistant doctors must attend teaching activities every 31 time steps. \\
        $\forall a, t (t \mod 31 = 0 \land Role(a, Chief \lor Assistant) \implies Attends(a, TeachingActivity, t))$ \\ \bottomrule
    \end{tabular}
\end{table}

\begin{figure}[h]
  \centering
  \includegraphics[width=0.75\linewidth]{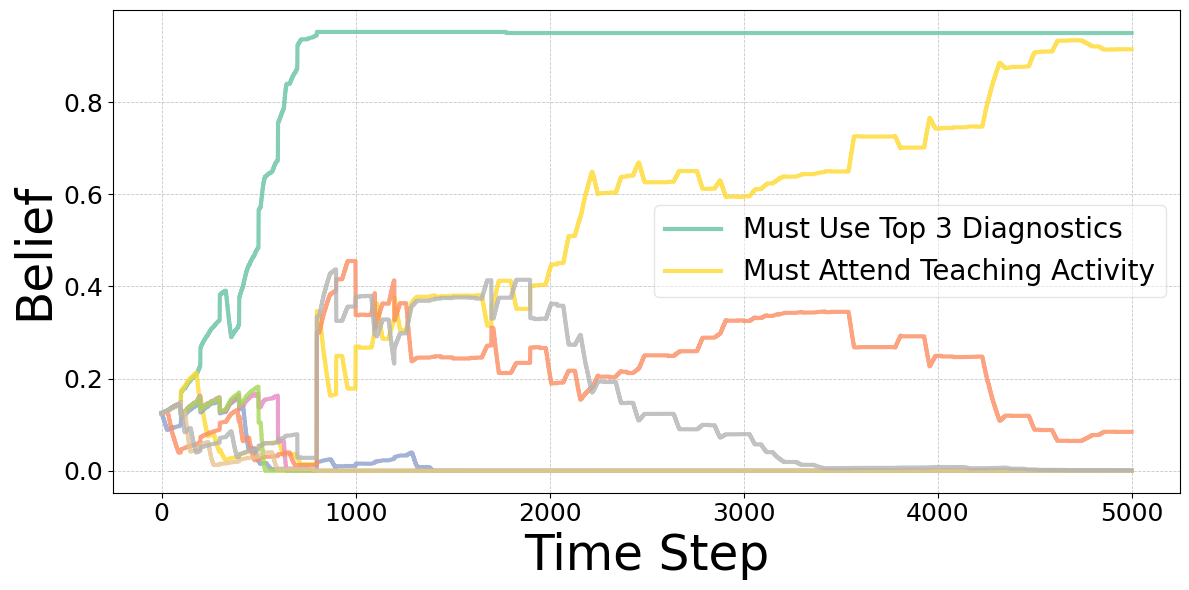}
  \caption{Belief Dynamics of Core and Other Norms}
  \label{fig:Belief_Dynamics_Core_Other_Norms}
  % \Description{A line chart showing the belief dynamics of core norms (must_use_top_3_diagnostics and must_attend_teaching_activity) and other norms over 2500 time steps. The core norms are highlighted in the legend, while other norms are represented by colored lines without individual annotations.}
\end{figure}
We extended the experimental time steps of the 20 agents to 5000. Figure~\ref{fig:Belief_Dynamics_Core_Other_Norms} illustrates the belief dynamics of agent 0 across all norms within the Markov game environment during this period. It can be observed that the agent gradually learns the two medical protocols. We computed the average beliefs for all agents across the norms at this stage and sorted them in descending order. The protocol \texttt{must\_use\_top\_3\_diagnostics} with a value of 0.8667 and \texttt{must\_attend\_teaching\_activity} with a value of 0.8023 ranked first and second, respectively. See Table~\ref{tab:average_beliefs} for details. From Figure~\ref{fig:Belief_Dynamics_Core_Other_Norms}, it can be observed that during the initial 1000 steps, the belief value for the norm \texttt{must\_use\_top\_3\_diagnostics} generally exhibits a linear increase. However, the belief value for the norm \texttt{must\_attend\_teaching\_activity} remains mixed with other control norms and fails to form a clear belief initially. Through the agent's practice-based validation of norms, it can be observed that after completely violating the "must attend teaching activity" norm, the agent's belief in this norm shows a slight sudden increase. However, this process is also accompanied by sudden increases in belief for two other random control norms. By step 2000, the agent's belief in \texttt{must\_attend\_teaching\_activity} gradually diverges from the control norms, forming a more distinct belief trend. The experimental results exhibit the same trend as the learning of prescriptive norms by humans (natural agents) in real-world scenarios.

% \section{CONCLUSION}
% Our model validates that in distributed healthcare environments, collective descriptive norms, as demonstrated in the vertigo dataset, can be gradually formed through dynamic interaction processes. Our enhanced practice-driven Markov games further confirm the feasibility of autonomous agent systems learning strict medical norms in distributed dynamic interactions using approximately rational Bayesian rule induction. The two approaches in this model are proposed based on the nature of the medical norm system and its propagation channels, with their core focus on dynamic interactions within distributed healthcare scenarios.

\section{DISCUSSION AND FUTURE WORK}
Maintaining a higher level of awareness of descriptive norms in large-scale multi-agent systems (with far more than 80 medical agents) remains an interesting challenge. This mirrors real-world healthcare environments, where the expansion of the medical community from individual clinics to city-wide healthcare institutions, nationwide networks, and even international organizations presents deeper issues in designing large-scale medical scenarios \cite{Funkner2021CitywideQO}.

In the context of distributed healthcare, gaining beliefs about prescriptive norms through continuous interaction among agents in a Markov game environment is a novel approach. Deriving medical rules from actions and the data generated by those actions is inherently complex. While we have achieved belief updates in agents regarding correct norms by introducing external regulation mechanisms such as 'norm practice,' the deeper issue lies in exploring the general mathematical properties of the model’s internal probabilities. It is essential for a multi-agent system to autonomously and rationally infer correct and useful clinical protocols without relying on external regulation, ensuring that the system functions effectively and independently within the healthcare environment.

The diversity in types and dissemination channels of medical norms is vast, and this study focuses solely on clinical protocols and recommendations within the given dataset. Future research must explore the various challenges posed by the integration of medical norms and autonomous agent systems in broader healthcare scenarios.

\section*{Acknowledgments}
The research was supported by the Russian Science Foundation, agreement No. 24-11-00272, \url{https://rscf.ru/project/24-11-00272/}.

\printcredits

\bibliographystyle{model1-num-names}

\bibliography{cas-refs}


%%%%%%%%%%%%%%%%%%%%%%%%%%%%%%%%%%%%%%%%%%%%%%%%%%%%%%%%%%%%%%%%%%%%%%%%
\appendix
\clearpage % 确保附录从新的一页开始
\setcounter{page}{1} % 页码从 1 开始
\renewcommand{\thesection}{\Alph{section}} % 附录章节格式
\renewcommand{\thepage}{A-\arabic{page}} % 页码格式为 "A-1, A-2, ..."
\input{appendix} % 引用附录文件

\end{document}

%% file: appendix.tex
%%%%%%%%%%%%%%%%%%%%%%%%%%%%%%%%%%%%%%%%%%%%%%%%%%%%%%%%%%%%%%%%%%%%%%%%
%%% Adding Appendix Section
%%%%%%%%%%%%%%%%%%%%%%%%%%%%%%%%%%%%%%%%%%%%%%%%%%%%%%%%%%%%%%%%%%%%%%%%
% 在附录中使用

\clearpage
\setcounter{figure}{0}
\setcounter{table}{0}
\renewcommand{\thefigure}{\thesection.\arabic{figure}}
\renewcommand{\thetable}{\thesection.\arabic{table}}

\section{Descriptive norm-sharing experiments}
\label{appendix:additional_results}

% k-means -5 group
\begin{figure}[h]
  \centering
  \includegraphics[width=0.75\linewidth]{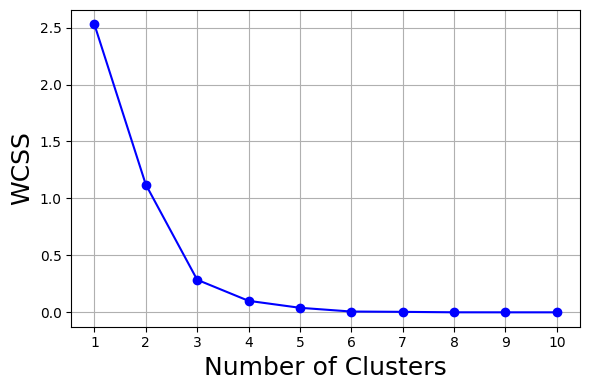}
  \caption{Elbow Method for Optimal Cluster Count}
  \label{fig:combined_clusters_elbow}
  % \Description{The Elbow Method plot showing WCSS values, where the curve levels off after K=5.}
\end{figure}

% 图1: 14个代理的收敛情况
\begin{figure}[h]
  \centering
  \includegraphics[width=0.75\linewidth]{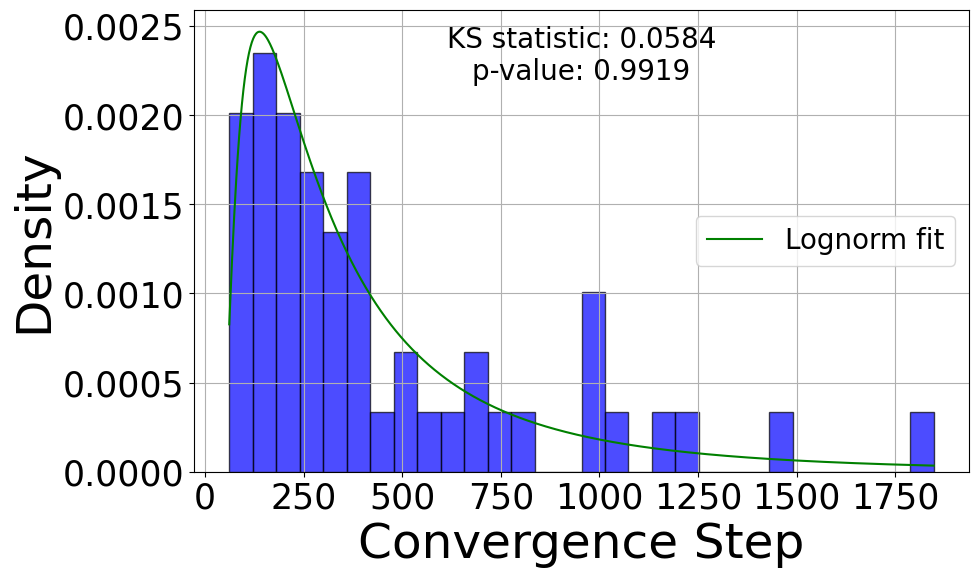}
  \caption{The convergence step distribution for 14 agents, fitted with a Lognormal distribution.}
  \label{fig:lognorm_14_agents}
  % \Description{Histogram of convergence steps for 14 agents with a fitted Lognormal distribution curve.}
\end{figure}

% 图3: 36个代理的收敛情况
\begin{figure}[h]
  \centering
  \includegraphics[width=0.75\linewidth]{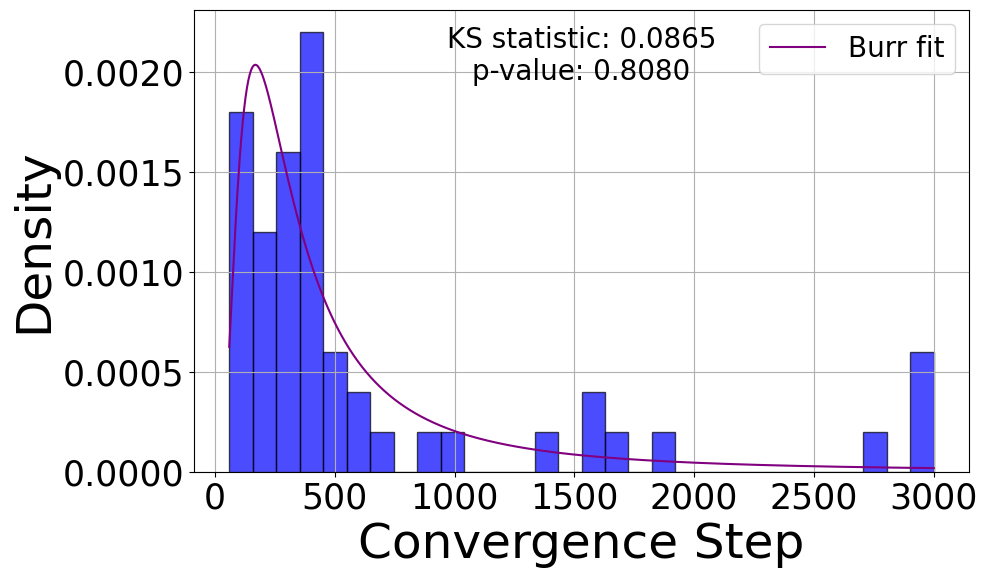}
  \caption{The convergence step distribution for 36 agents, fitted with a Burr distribution.}
  \label{fig:burr_36_agents}
  % \Description{Histogram of convergence steps for 36 agents with a fitted Burr distribution curve.}
\end{figure}

% 图4: 50个代理的收敛情况
\begin{figure}[h]
  \centering
  \includegraphics[width=0.75\linewidth]{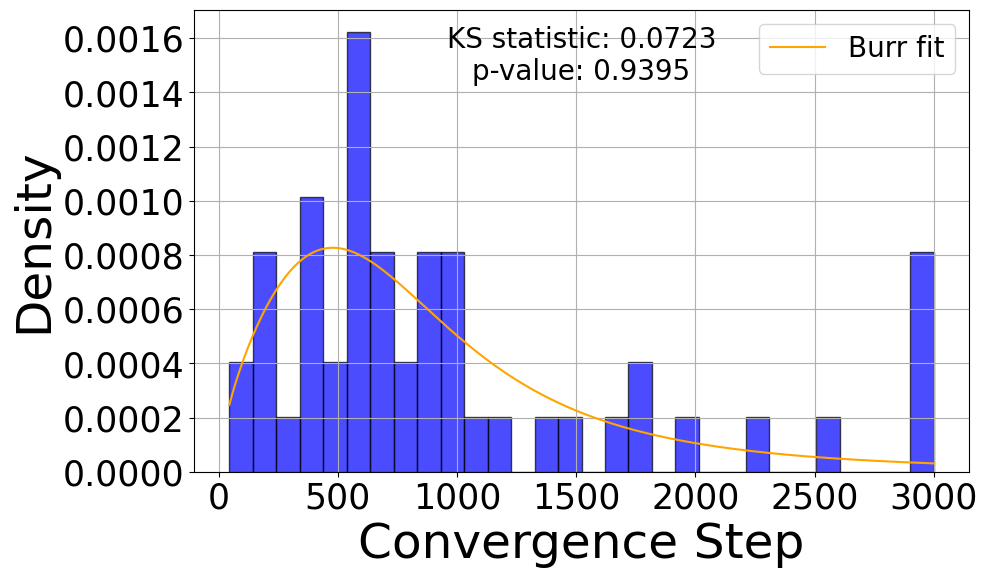}
  \caption{The convergence step distribution for 50 agents, fitted with a Burr distribution.}
  \label{fig:burr_50_agents}
  % \Description{Histogram of convergence steps for 50 agents with a fitted Burr distribution curve.}
\end{figure}

\begin{figure}[h]
  \centering
  \includegraphics[width=0.75\linewidth]{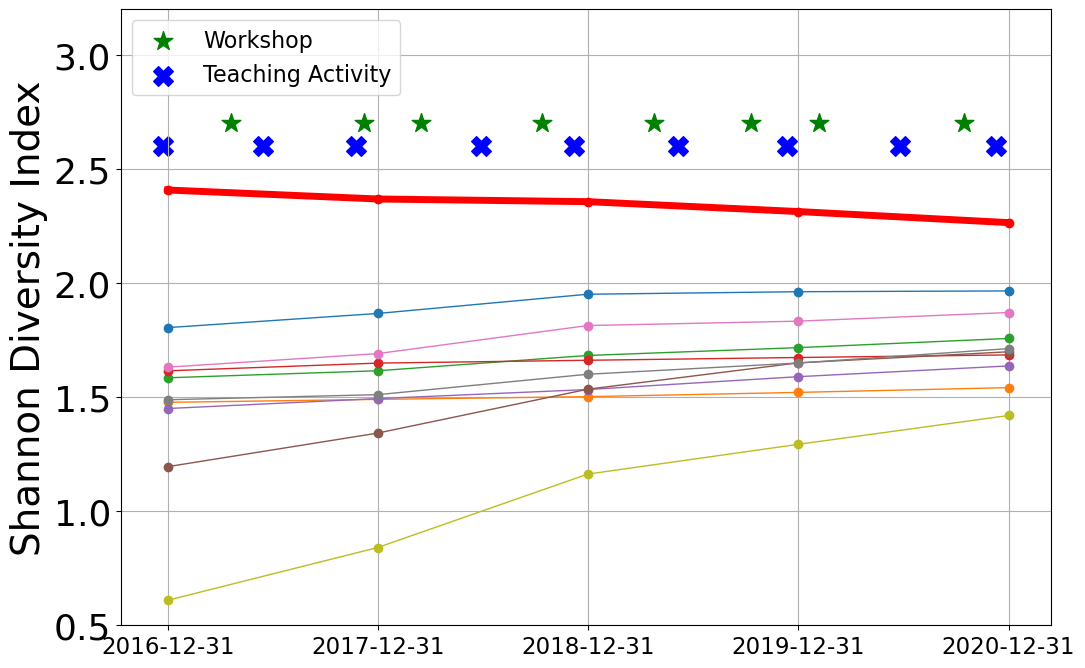}
  \caption{Shannon Diversity Index Trends for All Doctors (2016-2020).}
  \label{fig:shannon_diversity_all_doctors}
  % \Description{This figure shows the Shannon Diversity Index trends for all doctors from 2016 to 2020. The plot highlights the changes over time for each doctor, with a particular emphasis on Chief Physician (in red), showing a unique trend compared to others.}
\end{figure}

\begin{figure}[h]
  \centering
  \includegraphics[width=0.75\linewidth]{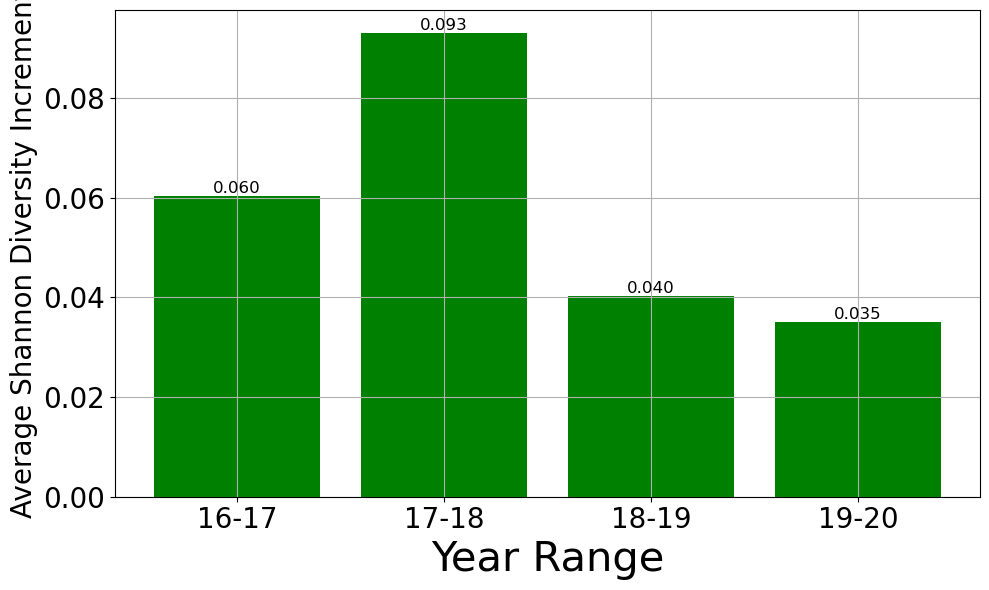}
  \caption{Average Yearly Shannon Diversity Index Increment.}
  \label{fig:average_shannon_increment}
  % \Description{This figure displays the increment of the average Shannon Diversity Index between consecutive years. Positive increments indicate an increase in diversity, while negative increments indicate a decrease, providing a clear visualization of biodiversity changes over time.}
\end{figure}

% 引用附录中的内容
Figure~\ref{fig:combined_clusters_elbow} shows the clustering results of K-means applied to the overall average practice Shannon Diversity Index of 10 doctors over 5 years.
Figure~\ref{fig:lognorm_14_agents} shows the convergence steps distribution for 14 agents, fitted with a Lognormal distribution. For 36 agents, Figure~\ref{fig:burr_36_agents} demonstrates the Burr distribution fit, and similarly, Figure~\ref{fig:burr_50_agents} illustrates the fit for 50 agents.

Figure~\ref{fig:shannon_diversity_all_doctors} shows the Practice Shannon Diversity Index for all doctors over a 5-year period, reflecting the dynamic development of practice beliefs in the clinic. The points represent key practice-sharing events, such as the first teaching activity on December 23rd, 2016, marked by a blue "X". The figure demonstrates the overall trend of belief convergence among the 10 doctors. The index for the senior chief physician (highlighted in bold red) shows a slight decline, while the indices for assistant doctors rise, indicating a dynamic evolution of collective medical practice that supports our model’s approach to norm perception and interaction.

Figure~\ref{fig:average_shannon_increment} presents the average yearly increment of the Shannon Diversity Index over multiple years. The average value of the Shannon Diversity Index of the practices of 10 neurologists from December 31 of each year to December 31 of the following year is calculated. For example, the first column represents the value of the Shannon Diversity Index from December 31, 2016 to December 31, 2017.

% 附录内容
\onecolumn
\setcounter{figure}{0}
\setcounter{table}{0}
\renewcommand{\thefigure}{\thesection.\arabic{figure}}
\renewcommand{\thetable}{\thesection.\arabic{table}}
\renewcommand{\theequation}{\thesection.\arabic{equation}}

\section{Norm Practice and Smooth Updates}
\label{appendix:Norm Practice and Smooth Updates}
Assumption 1 (Section 2.4) establishes the integration of the norm verification and adjustment process into agents' belief updating mechanisms. This process includes an external regulation mechanism that refines agents' beliefs based on the observed effectiveness of norms in the environment. After identifying the most relevant norms within the norm space $\mathcal{H}$, agents adjust their beliefs $b(n_m)$ about each norm $n_m \in \mathcal{H}$ according to its observed efficacy in practice. This adjustment modifies the belief update rule, where the beliefs are reweighted by a regulation factor $\alpha \in [0,1]$, scaling the belief values based on the empirical success of the norm: 
\begin{eqnarray}\label{eq:belief_update} b_i'(n_m) & = & \alpha \cdot b_i(n_m) \end{eqnarray} 
The adjustment factor $\alpha$ is determined through an external validation process that evaluates the effectiveness of norms in guiding appropriate actions. Once adjusted, beliefs are normalized to ensure they form a valid probability distribution: % 
\begin{eqnarray}\label{eq:belief_normalization} b_i''(n_m) & = & \frac{b_i'(n_m)}{\sum_{n_m \in \mathcal{H}} b_i'(n_m)} \end{eqnarray} 
This external regulation mechanism ensures that agents do not converge to non-functional norms by continuously adapting their belief structures based on practical outcomes in the environment.

In addition to this external regulation through "practice verification," adaptive learning rates and momentum mechanisms were introduced into the process of agents’ concept updates.

Specifically, a momentum mechanism was implemented to smooth gradient updates during the belief update process for prescriptive norms, mitigating the effects of noise and unstable gradients. The momentum update is defined as: 
\begin{eqnarray}\label{eq:momentum_update} m_{i}^{t+1}(n_m) & = & \beta_1 \cdot m_i^t(n_m) + (1 - \beta_1) \cdot \nabla \mathcal{L}i(n_m) \end{eqnarray} 
where $m{i}^{t+1}(n_m)$ represents the momentum for norm $n_m$ at step $t+1$, $\nabla \mathcal{L}_i(n_m)$ is the current gradient for the norm, and $\beta_1$ is the momentum coefficient.

An adaptive learning rate was also introduced, dynamically adjusting the step size during updates to ensure stability in the learning process. The adaptive learning rate is defined as: % 
\begin{eqnarray}\label{eq:adaptive_lr} \alpha_{i}^{t+1}(n_m) & = & \frac{\alpha}{\sqrt{v_{i}^{t+1}(n_m)} + \epsilon} \end{eqnarray} 
where $\alpha_{i}^{t+1}(n_m)$ is the adaptive learning rate, $\alpha$ is the initial learning rate, $v_{i}^{t+1}(n_m)$ is the second moment estimate for norm $n_m$, and $\epsilon$ is a small constant to prevent division by zero.

To support this mechanism, the second moment estimate $v_{i}^{t+1}(n_m)$ was updated dynamically as: 
\begin{eqnarray}\label{eq:second_moment} v_{i}^{t+1}(n_m) & = & \beta_2 \cdot v_{i}^{t}(n_m) + (1 - \beta_2) \cdot \left( \nabla \mathcal{L}{i}(n_m) \right)^{2} \end{eqnarray} 
where $v{i}^{t+1}(n_m)$ represents the second moment estimate for norm $n_m$ at step $t+1$, and $\beta_2$ is the RMSprop coefficient.

\onecolumn
\setcounter{figure}{0}
\setcounter{table}{0}
\renewcommand{\thefigure}{\thesection.\arabic{figure}}
\renewcommand{\thetable}{\thesection.\arabic{table}}

\section{Prescriptive Norm Induction Experiments}
\label{appendix:Prescriptive Norm Induction}

\begin{figure}[h]
  \centering
  \includegraphics[width=0.8\textwidth]{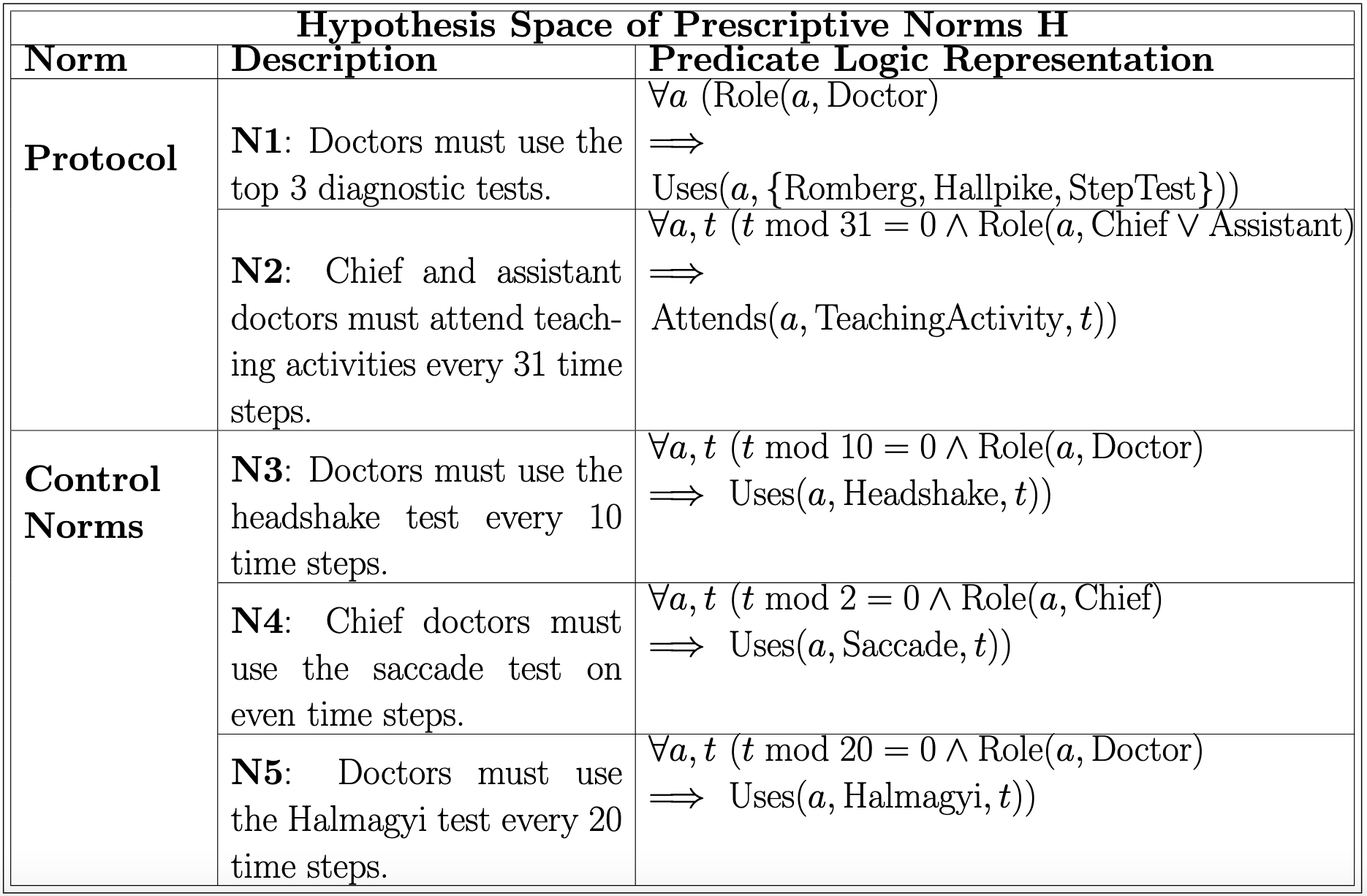}
  \caption{\hspace{0.5cm} Hypothesis Space Representation during Model Training}
  \label{fig:Hypothesis_Space_H}
  % \Description{A diagram showing the hypothesis space with nodes representing different hypothesis classes and transitions that occur as agents adapt to new medical information.}
\end{figure}

Figure~\ref{fig:Hypothesis_Space_H} presents the core protocols $\mathcal{N}_1$ and $\mathcal{N}_2$, along with three randomly selected control norms. In each experiment, we use two core protocols and six control norms as the target norms for that particular run. This represents part of the overall structure of the Hypothesis Space of Prescriptive Norms $\mathcal{H}$.

Table~\ref{tab:average_beliefs} presents the eight norms that agents in the Hypothesis Space of Prescriptive Norms $\mathcal{H}$ in predicate logic were able to learn most effectively. Among these, the core clinical protocols exhibit the highest belief values, while the non-core norms display lower beliefs. This indicates that agents have avoided the high probability of learning a large number of "silly" norms. With a certain level of external regulation, agents were able to develop an understanding of and increase their belief in medical norms.

\begin{table}[h]
  \caption{\hspace{0.5cm} Average Beliefs for Each Norm in Descending Order}
  \label{tab:average_beliefs}
  \centering
  \begin{tabular}{|p{0.7\textwidth}|c|}
    \hline
    \textit{Norm} & \textit{Average Belief} \\ \hline  % 统一使用 \hline
		\texttt{must\_use\_top\_3\_diagnostics} & 0.8667 \\ 
		\texttt{must\_attend\_teaching\_activity} & 0.8023 \\
		\texttt{must\_use\_babinskogo\_veylia\_if\_not\_used\_recently} & 0.2989 \\
		\texttt{must\_use\_romberga\_if\_not\_used\_recently} & 0.2450 \\
		\texttt{must\_use\_headshake\_every\_10\_steps} & 0.2274 \\
		\texttt{must\_use\_shagovyi\_if\_not\_used\_last\_time} & 0.1138 \\
		\texttt{must\_use\_khalmagi\_every\_20\_steps} & 0.1046 \\
		\texttt{must\_use\_sakkad\_if\_even\_step} & 0.0320 \\ \hline  % 使用统一的 \hline
	\end{tabular}
\end{table}

\clearpage
\setcounter{figure}{0}
\setcounter{table}{0}
\renewcommand{\thefigure}{\thesection.\arabic{figure}}
\renewcommand{\thetable}{\thesection.\arabic{table}}
\section{prescriptive control norms}
\label{appendix:control norms}

The following is a list of 30 control norms, arranged in ascending order of logical complexity based on their conditions:

\begin{enumerate}
    \item \textbf{If the time step is even} $\rightarrow$ Assistant doctors must use the Step test.
    
    \item \textbf{If the time step is even} $\rightarrow$ Chief doctors must use the Saccade test.
    
    \item \textbf{If the time step is odd} $\rightarrow$ Chief doctors must use the Babinski-Weil test.
    
    \item \textbf{If the time step is divisible by 4} $\rightarrow$ Assistant doctors must use the Hallpike test.
    
    \item \textbf{If the time step is divisible by 5} $\rightarrow$ Doctors must use the Saccade test.
    
    \item \textbf{If the time step is divisible by 6} $\rightarrow$ Assistant doctors must use the Romberg test.
    
    \item \textbf{If the time step is divisible by 7} $\rightarrow$ Chief doctors must use the Halmagyi test.
    
    \item \textbf{If the time step is divisible by 9} $\rightarrow$ Chief doctors must use the Babinski-Weil test.
    
    \item \textbf{If the time step is divisible by 10} $\rightarrow$ Doctors must use the Headshake test.
    
    \item \textbf{If the time step is divisible by 13} $\rightarrow$ Doctors must use the Romberg test.
    
    \item \textbf{If the time step is divisible by 15} $\rightarrow$ Doctors must use the Hallpike test.
    
    \item \textbf{If the time step is divisible by 20} $\rightarrow$ Doctors must use the Halmagyi test.
    
    \item \textbf{If the time step is divisible by 25} $\rightarrow$ Assistant doctors must use the Halmagyi test.
    
    \item \textbf{If the time step is a prime number} $\rightarrow$ Chief doctors must use the Step test.
    
    \item \textbf{If doctors didn't use the Headshake test in the previous diagnosis} $\rightarrow$ They must use the Headshake test now.
    
    \item \textbf{If assistant doctors didn't use the Saccade test in the last diagnosis} $\rightarrow$ They must use the Saccade test now.
    
    \item \textbf{If chief doctors didn't use the Hallpike test in the previous diagnosis} $\rightarrow$ They must use the Hallpike test now.
    
    \item \textbf{If chief doctors didn't use the Step test in the previous diagnosis} $\rightarrow$ They must use the Step test now.
    
    \item \textbf{If assistant doctors haven't used the Step test in the last 3 time steps} $\rightarrow$ They must use the Step test now.
    
    \item \textbf{If assistant doctors haven't used the Headshake test in the last 4 time steps} $\rightarrow$ They must use the Headshake test now.
    
    \item \textbf{If doctors haven't used the Hallpike test in the last 5 time steps} $\rightarrow$ They must use the Step test now.
    
    \item \textbf{If assistant doctors haven't used the Babinski-Weil test in the last 5 time steps} $\rightarrow$ They must use the Babinski-Weil test now.
    
    \item \textbf{If chief doctors haven't used the Romberg test in the last 7 time steps} $\rightarrow$ They must use the Romberg test now.
    
    \item \textbf{If doctors haven't used the Babinski-Weil test in the last 8 time steps} $\rightarrow$ They must use the Babinski-Weil test now.
    
    \item \textbf{If doctors haven't used the Romberg test in the last 10 time steps} $\rightarrow$ They must use the Romberg test now.
    
    \item \textbf{If doctors haven't used the Headshake test in the last 12 time steps} $\rightarrow$ They must use the Headshake test now.
    
    \item \textbf{If doctors used more than 4 tests in the last diagnosis} $\rightarrow$ They must use the Romberg test now.
    
    \item \textbf{If assistant doctors used fewer than 3 diagnostics in the last diagnosis} $\rightarrow$ They must use the Babinski-Weil test now.
    
    \item \textbf{If doctors didn't use any tests in the last 2 time steps} $\rightarrow$ They must use the Saccade test now.
    
    \item \textbf{If the time step is divisible by 14} $\rightarrow$ Doctors must use at least two of the following tests: Headshake, Saccade, Halmagyi.
\end{enumerate}